\documentclass{article} % For LaTeX2e
\usepackage{arxiv,times}
\usepackage{enumitem}
\usepackage{amsmath,amsfonts,bm}

\def\eqref#1{equation~\ref{#1}}
\def\1{\bm{1}}

\DeclareMathAlphabet{\mathsfit}{\encodingdefault}{\sfdefault}{m}{sl}
\SetMathAlphabet{\mathsfit}{bold}{\encodingdefault}{\sfdefault}{bx}{n}

\usepackage{subcaption}
\usepackage{colortbl}
\usepackage{booktabs}
\usepackage{graphicx}
\usepackage[table]{xcolor}
\usepackage{hyperref}
\usepackage{url}
\usepackage{hyperref}
\usepackage{graphicx}
\usepackage{enumitem}
\usepackage{booktabs}
\usepackage{multirow}
\usepackage{array}
\usepackage{subcaption}
\usepackage{amssymb}
\usepackage{amsthm}
\usepackage{tcolorbox}
\usepackage{algorithm}
\usepackage{algpseudocode}
\usepackage{float}
\usepackage{listings}
\definecolor{groupgray}{RGB}{244,245,247}
\definecolor{sgblue}{RGB}{238,246,252}

\usepackage{listings}
\usepackage{xcolor}
\usepackage{tcolorbox}
\tcbuselibrary{listings,skins,breakable}

\definecolor{sgtitle}{RGB}{72,84,96}
\definecolor{sgback}{RGB}{247,248,250}
\definecolor{sgframe}{RGB}{120,128,136}

\newtcblisting{sglisting}[2][]{
  enhanced,
  breakable,
  listing only,
  colback=sgback,
  colframe=sgframe,
  colbacktitle=sgtitle,
  coltitle=white,
  fonttitle=\bfseries\small,
  title={#2},
  boxrule=0.5pt,
  arc=1.5pt,
  outer arc=1.5pt,
  left=1.0mm,
  right=1.0mm,
  top=0.8mm,
  bottom=0.8mm,
  titlerule=0pt,
  listing options={
    basicstyle=\ttfamily\scriptsize,
    columns=fullflexible,
    keepspaces=true,
    breaklines=true,
    showstringspaces=false,
    aboveskip=0pt,
    belowskip=0pt,
    lineskip=-0.5pt
  },
  #1
}

\title{StateGuard: Analytical-State Management with Validity-Aware Intervention for Long-Horizon Data Agents}

\author{
Wenle Liao, Zhao Wang, Jingchao Zhang, Jiajie Jin, Yimeng Xu
, Zhicheng Dou\thanks{Corresponding author.} \\
Gaoling School of Artificial Intelligence \\
Renmin University of China
}

\iclrfinalcopy % Uncomment for camera-ready version, but NOT for submission.
\begin{document}

\maketitle

\begin{abstract}
LLM-based agents have shown strong capabilities in automated data analysis and are increasingly moving toward long-horizon, multi-stage analytical workflows. However, as the analytical process evolves, constraints, variables, and conclusions remain implicitly embedded in interaction histories, making it difficult for agents to track which analytical artifacts remain valid over increasingly long horizons and changing dependencies. Consequently, stale artifacts may be silently inherited, propagating errors to downstream stages. To address this challenge, we propose \textbf{StateGuard}, an analytical-state validity management framework for long-horizon data agents. StateGuard externalizes evolving analytical progress into a state graph containing constraints, versioned variables, intermediate conclusions, and cross-state relations, treating each state as an executable, verifiable, and traceable object rather than textual memory alone. StateGuard maintains state validity through evidence-grounded verification and hierarchical intervention. To equip StateGuard with these capabilities, we first introduce \emph{Manager-Oriented Counterfactual Supervision}, which constructs 3K state-centric trajectories through counterfactual runtime synthesis to fine-tune StateGuard for state maintenance, verification, and repair. We then apply \emph{Validity-Guided Policy Optimization}, using runtime validity evidence to provide fine-grained learning signals for protocol correctness, state grounding, and intervention quality. Experiments on three diverse long-horizon data-analysis benchmarks show that StateGuard consistently improves data-agent performance while reducing dependency-induced downstream error propagation, demonstrating the advantages of explicit analytical-state management for reliable long-horizon data analysis.
\end{abstract}

\section{Introduction}

Large Language Model (LLM)-based agents have recently shown strong capabilities in automated data analysis, combining reasoning, code execution, and tool use to solve increasingly complex data-science tasks~\citep{yao2023reactsynergizingreasoningacting, wang2024executablecodeactionselicit}. Recent data agents further extend this paradigm toward more autonomous and multi-stage analytical workflows through dynamic task decomposition, iterative verification, and agentic training~\citep{hong-etal-2025-data, nam2026dsstardatascienceagent, zhang2025deepanalyzeagenticlargelanguage, qiao2026scalinggeneralistdataanalyticagents}. As these systems move beyond isolated queries toward long-horizon analysis, agents must repeatedly inspect data, revise intermediate results, and synthesize conclusions across dependent stages~\citep{lei2025dacompbenchmarkingdataagents, xu2026longdsbenchfailurelonghorizonagentic}. However, recent evaluations show
that agent performance degrades substantially as the analytical process evolves and dependencies extend over longer horizons, where early inconsistencies can be inherited by later computations and propagate through the analytical process, ultimately leading to an incorrect result~\citep{xu2026longdsbenchfailurelonghorizonagentic}.

\begin{figure}[t]
    \centering
    \includegraphics[width=\linewidth]{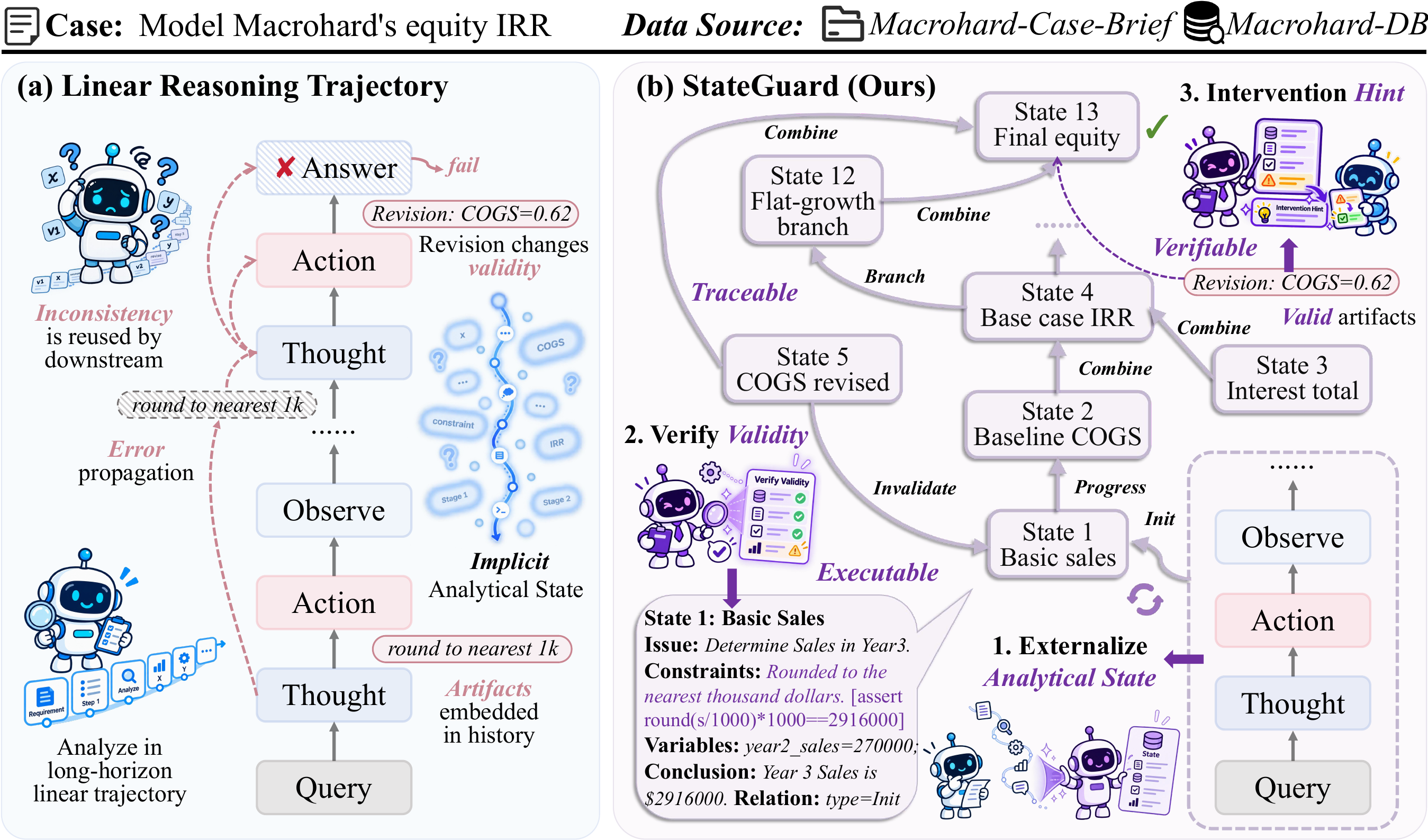}
    \caption{Comparison between our proposed StateGuard and linear reasoning trajectory.}
    \label{fig:1}
\end{figure}

To address this challenge, recent work has improved both long-horizon execution and context management. Data-science agents structure complex workflows through task decomposition and planning, while iterative process verifications inspect intermediate execution and detect errors~\citep{hong-etal-2025-data,nam2026dsstardatascienceagent,10.1145/3770855.3819049}. In parallel, memory-based approaches compress, retrieve, or organize interaction histories according to the agent's execution progress, helping preserve useful information across distant steps~\citep{chen2026semanticorganizationmemoryexecution,hu2026samstateadaptivememorylonghorizon}. However, these methods mainly operate over tasks, individual steps, or generic contexts rather than explicitly maintaining the validity of derived artifacts as dependencies evolve. As illustrated in Figure \ref{fig:1}, a previous constraint such as \emph{round to nearest 1k} may remain in the context without tracking its validity conditions, staying executable yet inconsistent with later revisions. Such inconsistencies often remain undetected and can silently propagate further, leading to cascading failures that are difficult to localize.

Our key observation is that analytical states in data agent tasks differ fundamentally from generic textual memory or execution progress. Data analysis is naturally organized around formalizable constraints, executable computations, and variables. Their validity is directly testable, and dependencies can be explicitly traced. An analytical state can therefore be treated not merely as information to be remembered, but as an executable, verifiable, and dependency-traceable object. Its validity can be checked with runtime evidence, while its downstream influence
can be traced through state and variable dependencies, as the \emph{revised COGS value} invalidates previous artifacts. Motivated by this observation, we propose \textbf{StateGuard}, which externalizes evolving analytical progress into structured states and maintains their validity while detecting potential inconsistencies during execution.

StateGuard operates as an external management layer over the underlying data-analysis process, maintaining explicit analytical states throughout execution. Each state records the current issue, constraints, versioned variables, intermediate conclusions, and relations to prior states. Newly formed states are verified against executable runtime evidence, while relevant verified states are surfaced to the agent as compact state hints. When an inconsistency is detected, StateGuard traces state and variable dependencies to localize the violated variable or constraint, and performs evidence-grounded hierarchical repair, including stronger artifact-level intervention when necessary. This makes the analytical process explicit, verifiable, and traceable throughout long-horizon execution.

To further improve StateGuard's state maintenance, verification, and intervention capabilities, we design a two-stage training pipeline. We first introduce \textbf{Manager-Oriented Counterfactual Supervision}, which combines relation-based task synthesis with counterfactual intervention for error injection to generate 3K manager-oriented trajectories for supervised fine-tuning, teaching StateGuard the protocols for state maintenance, verification, and intervention. We then introduce \textbf{Validity-Guided Policy Optimization}, which turns runtime validity evidence into fine-grained signals over protocol correctness, state grounding, and intervention quality, beyond outcome-only supervision.

% Training StateGuard directly with outcome-level reinforcement learning is difficult because it must first learn the structured protocol of state maintenance, verification, and intervention. We therefore construct and release 3K high-quality manager-oriented trajectories through a state-centric synthesis pipeline that combines relation-graph-based task construction, counterfactual intervention, and executable runtime verification. We first perform supervised fine-tuning to teach StateGuard runtime protocol and lifecycle, and then further optimize the policy with DAPO using validity evidence runtime rewards for protocol correctness, state grounding, and intervention quality.

We evaluate StateGuard on three diverse and challenging long-horizon data-analysis benchmarks, including LongDS-Bench~\citep{xu2026longdsbenchfailurelonghorizonagentic}, DABstep~\citep{egg2025dabstepdataagentbenchmark}, and DAComp-DE~\citep{lei2025dacompbenchmarkingdataagents}. Across different agents and task settings, StateGuard consistently improves long-horizon performance and achieves the best results across multiple evaluations. We further evaluate dependency error propagation with \emph{Dependency Contamination Rate (DCR)}, observing that StateGuard reduces downstream contamination from incorrect upstream. 

Our key contributions are summarized as follows:
\begin{itemize}[
    leftmargin=1.5em,
    itemsep=2pt,
    parsep=0pt,
    topsep=0pt,
    partopsep=0pt
]   
    \item \textbf{Executable analytical-state externalization for long-horizon data analysis.}
    We represent long-horizon data analysis as explicit analytical states with constraints, versioned variables, conclusions, and cross-state relations, making their validity verifiable and dependencies traceable.

    \item \textbf{Runtime state validity management and evidence-grounded intervention.}
    We propose StateGuard, which maintains and verifies analytical states and artifacts during execution, and performs validity evidence verification and hierarchical intervention before errors propagate further.

    \item \textbf{Manager-oriented counterfactual supervision and validity-guided policy optimization.}
    We construct 3K manager-oriented trajectories via relation-based task construction and counterfactual runtime synthesis, and further optimize StateGuard with validity evidence over protocol, grounding, and intervention. Experiments across three benchmarks show consistent gains and reduced dependency error propagation.
\end{itemize}

\section{Related Work}

\paragraph{LLM Agents for Data Science and Long-Horizon Data Analysis.}
LLM-based agents are increasingly used for automated data science through reasoning, code execution, tool use, and database interaction~\citep{yao2023reactsynergizingreasoningacting,wang2024executablecodeactionselicit}. Recent systems further extend end-to-end data science workflows through task decomposition, iterative planning and verification, and agentic training~\citep{nam2026dsstardatascienceagent,zhang2025deepanalyzeagenticlargelanguage,10.1145/3770855.3818002}. Benchmarks such as LongDS and DAComp show that long-range dependencies, evolving requirements, and cross-stage consistency remain major challenges~\citep{xu2026longdsbenchfailurelonghorizonagentic,lei2025dacompbenchmarkingdataagents}. Meanwhile, structured data-analytic frameworks have improved the task organization~\citep{hong-etal-2025-data}, but evolving analytical states and artifacts remain largely implicit in the interaction trajectory. StateGuard instead explicitly maintains evolving analytical states and their cross-stage dependencies during execution.

\paragraph{Agent Memory and Context Management.}
Prior work improves long-horizon agents through external memory, context compression, and learned memory operations~\citep{packer2024memgptllmsoperatingsystems,xu2025amemagenticmemoryllm}. More recent approaches adapt memory access to the agent's process or organize execution over long interactions~\citep{hu2026samstateadaptivememorylonghorizon,chen2026semanticorganizationmemoryexecution}. However, they mainly address what information should be stored, retrieved, or summarized, rather than whether a previously derived analytical state is still valid. In data-analysis tasks, artifacts may remain in memory even after the conditions on which they depend have changed. StateGuard instead treats analytical state as an executable, verifiable, and dependency-traceable object rather than textual memory alone, focusing on the validity of previously derived artifacts as the analytical process evolves. 

\paragraph{Process Verification and Failure Repair.}
Recent work studies process-level verification to evaluate intermediate reasoning and detect silent errors~\citep{10.1145/3770855.3819049}. Failure-attribution methods further identify the action or trajectory step responsible for unsuccessful outcomes~\citep{barke2026agentrxdiagnosingaiagent,zhang2025agentracerinducingfailurellm,zhang2025agentcausestaskfailures}, while online auditing attempts to detect failures before task completes~\citep{zhang2026agentforesightonlineauditingearly}. Other approaches repair failed trajectories through reflection, replay, or rollback~\citep{hao2026speculativerollbackcorrectionqualitydiverse}. However, these methods mainly supervise trajectory or failure at individual steps or generic trajectory positions. StateGuard instead models inconsistencies as invalid states or variables, traces how they propagate through dependencies, performs evidence-grounded intervention and hierarchical repair over analytical-state integrity.

\section{Methodology}

\subsection{Problem Formulation}
We consider a long-horizon data-analysis task \(\mathcal{T}=(q,\mathcal{D})\), where \(q\) is the user query and \(\mathcal{D}\) denotes the accessible data sources. A data-analysis agent \(\pi\) interacts with the execution environment through iterative reasoning, actions, and observations, producing a trajectory
\begin{equation}
    \label{eq:trajectory}
    \tau=\{(r_t,a_t,o_t)\}_{t=1}^{T},
\end{equation}
where \(r_t\), \(a_t\), and \(o_t\) denote the reasoning, action, and resulting observation at step \(t\).

Long-horizon analysis continuously produces intermediate results that may be reused, revised, invalidated, or combined. We represent the evolving analytical progress as a sequence of states \(\mathcal{S}=\{S_1,\ldots,S_k\ldots,S_K\}\), where
\begin{equation}
\label{eq:analytical_state}
\qquad
S_k=(I_k,C_k,V_k,N_k,R_k).
\end{equation}
Here, \(I_k\) denotes the current analytical issue, \(C_k\) the associated constraints, \(V_k\) the used variables and their versions, \(N_k\) the intermediate conclusions, and \(R_k\) the semantic relations to previous states. These relations induce a state relation graph that captures how the current analysis depends on or evolves from earlier states, enabling related states to be explicitly retrieved and traced when needed.

Rather than recovering context repeatedly from evolving analytical progress, our goal is to maintain the validity and consistency of analytical states throughout execution. Specifically, StateGuard externalizes evolving analytical progress into structured states, verifies whether newly formed states remain valid under their current constraints and cross-state dependencies, and intervenes before inconsistencies propagate to downstream analysis.
\subsection{Overview of StateGuard}

\begin{figure}[t]
    \centering
    \includegraphics[width=\linewidth]{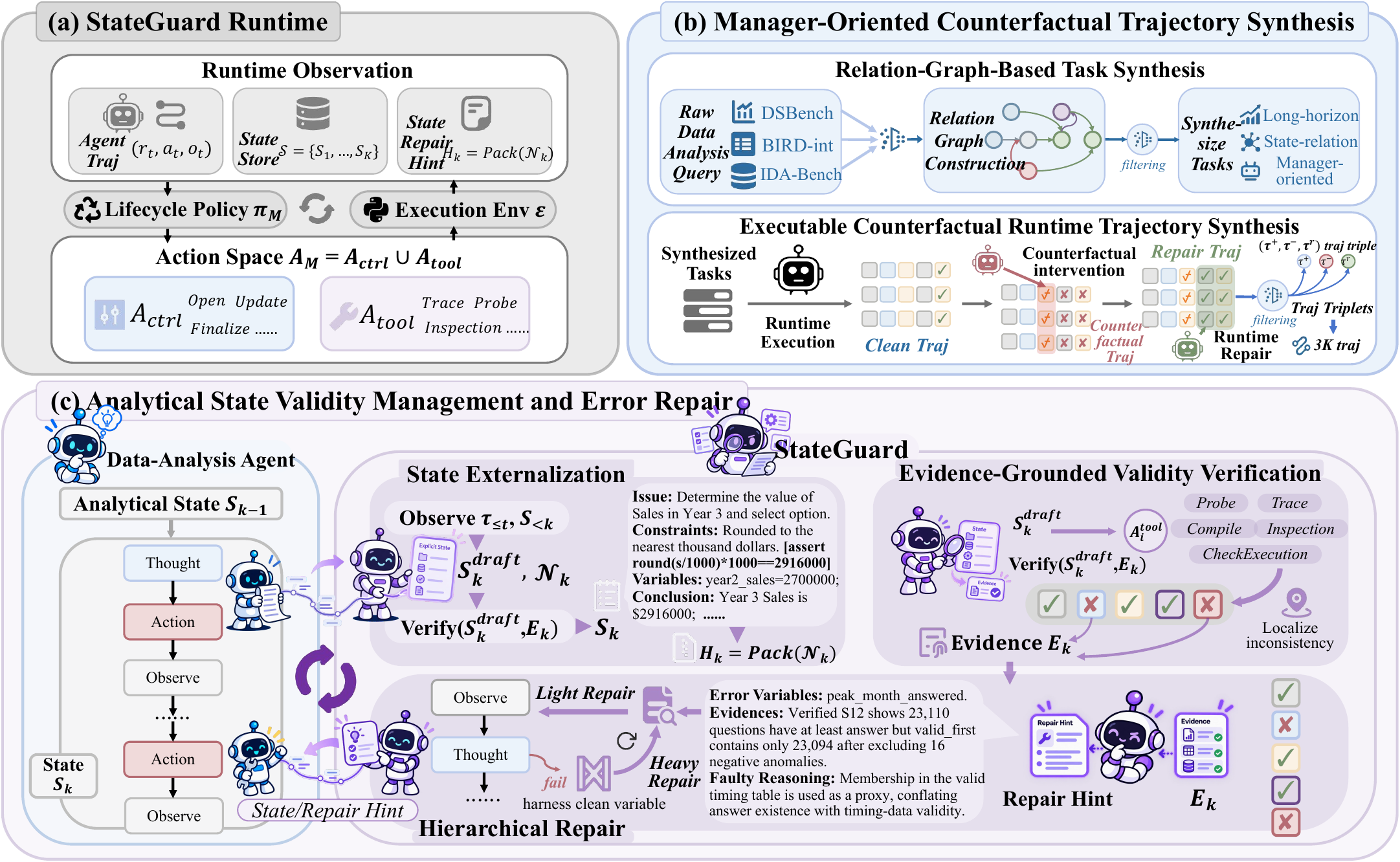}
    \caption{Overview of the StateGuard framework. StateGuard externalizes analytical states during agent execution, performs evidence-grounded validity verification and hierarchical repair, learns from manager-oriented trajectories constructed via relation-based and counterfactual synthesis.}
    \label{fig:2}
\end{figure}

As illustrated in Figure \ref{fig:2}, StateGuard augments a standard data-analysis agent with an external runtime manager that maintains explicit analytical states throughout execution. Unlike multi-agent systems, the data-analysis agent is the only task executor, while StateGuard only provides hints through a restricted state interface without participating in task planning or answer generation. Given the current agent trajectory \(\tau_{\le t}\) and previously committed states \(\mathcal{S}_{<k}\), StateGuard constructs and verifies the next analytical state before incorporating it into the persistent state store.

We organize the StateGuard action space $\mathcal{A}_M$ into two categories: control actions $\mathcal{A}_{\mathrm{ctrl}}$ and tool actions $\mathcal{A}_{\mathrm{tool}}$. Control actions govern the analytical-state lifecycle and tool actions execute and provide evidence for state verification and error localization. We denote the control actions as
$\mathcal{A}_{\mathrm{ctrl}}=
\{\textsc{Open},\textsc{Update},\textsc{Finalize},\textsc{Repair},\textsc{Commit},\textsc{Resume},\textsc{Abandon},\textsc{Abstain}\}$,
and the tool actions as
$\mathcal{A}_{\mathrm{tool}}=
\{\textsc{CheckExecution},\textsc{Inspection},\textsc{Compile},\textsc{Probe},\textsc{Trace}\}$.

For state \(S_k\), StateGuard first opens a state draft $S_k^{\mathrm{draft}}$ and updates it from the observed analytical process. It may invoke tool actions \(a^{\mathrm{tool}}\in\mathcal{A}_{\mathrm{tool}}\) to obtain verification evidence \(E_k\), and samples the next control action as \(a^{\mathrm{ctrl}}\sim\pi_M(\cdot\mid S_k^{\mathrm{draft}},\mathcal{S}_{<k},E_k)\).

The overall state lifecycle can be summarized as
\[
\textsc{Open}
\rightarrow
\textsc{Update}
\xrightarrow[\text{validity verification}]
{\quad\text{evidence-grounded}\quad}
\begin{cases}
\textsc{Finalize}\rightarrow\textsc{Commit}, & \text{if consistent},\\
\textsc{Repair}\rightarrow\textsc{Update}, & \text{otherwise}.
\end{cases}
\]
StateGuard may also abstain when the current trajectory does not support a reliable state decision; otherwise, it resumes the agent for continued execution. The harness executes StateGuard decisions and enforces valid state transitions, while StateGuard itself only performs state reasoning, decision-making and verification during runtime. See details in Appendix~\ref{app:lifecycle_relation} and~\ref{app:actions}.

\subsection{Analytical State Validity Management and Error Repair}
As illustrated in Figure \ref{fig:2}(c), StateGuard maintains analytical states through a unified extraction, verification, and repair process. Rather than treating the raw interaction history as the state itself, StateGuard externalizes the evolving analytical process into structured states and continuously checks their validity against prior analytical context and executable evidence.

\paragraph{Analytical State Externalization.}
Given the observed agent trajectory \(\tau_{\le t}\) and previously committed states \(\mathcal{S}_{<k}\), StateGuard constructs a draft state
\begin{equation}
\label{eq:draft_state}
S_k^{\mathrm{draft}}
=
f_{\mathrm{state}}(\tau_{\le t},\mathcal{S}_{<k}),
\end{equation}
extracting the current analytical issue, constraints, versioned variables, intermediate conclusions, and relations. The relation \(R_k\) is selected from \{ \textsc{Init}, \textsc{Progress}, \textsc{Branch}, \textsc{Invalidate}, \textsc{Combine} \}, capturing how the current stage evolves from prior states and inducing a state relation graph for dependency tracing.

For prior states associated with \(S_k\), denoted \(\mathcal{N}_k=\{S_j\in\mathcal{S}_{<k}\mid S_j\text{ is related to }S_k\}\), where \(\mathcal{N}_k\) denotes the relevant prior states of \(S_k\). StateGuard packages relevant issues, variables, conclusions and relations into a compact verified state hint \(H_k=\mathrm{Pack}(\mathcal{N}_k)\), allowing the agent to access valid analytical context and avoid confusing incompatible intermediate artifacts.

\paragraph{Evidence-Grounded Validity Verification.}
StateGuard verifies each draft state against evidence \(E_k\) from the agent trajectory, related states \(\mathcal{N}_k\), and executable runtime.
Rather than applying a fixed check-list, it invokes a tool action
\(a^{\mathrm{tool}}\in\mathcal{A}_{\mathrm{tool}}\) according to the current state and evidence:
\begin{equation}
\label{eq:verification_evidence}
a^{\mathrm{tool}}
\sim
\pi_M\!\left(
\cdot
\mid
S_k^{\mathrm{draft}},
\mathcal{N}_k,
E_k
\right),
\qquad
E_k \leftarrow E_k \cup \{e\},
\end{equation}
where \(e\) denotes the evidence returned by invoked tool action. The tools provide complementary evidence through
\emph{state trace}, \emph{syntax compile}, \emph{code inspection},
\emph{execution check}, and \emph{independent probe}. StateGuard combines these evidences with reasoning to verify constraints, variables, conclusions, and cross-state consistency, while the harness translates StateGuard's decisions into executable checks and automatically enforces corresponding constraints and valid state transitions.

The final judgment is \(z_k=\mathrm{Verify}(S_k^{\mathrm{draft}},E_k)\in\{\mathrm{consistent},\mathrm{inconsistent}\}\). A consistent state proceeds to finalization and commitment; otherwise, StateGuard localizes the inconsistency to violated constraints or invalid variables, and traces the related states when necessary. StateGuard therefore complements rather than replaces data-agent underlying semantic
task-solving capability.

\paragraph{Hierarchical Repair.}
When an inconsistency is detected, let \(C_k^{\mathrm{viol}}\), \(V_k^{\mathrm{err}}\), and \(F_k\)
denote violated constraints, invalid variables, and faulty reasoning, respectively.
StateGuard summarizes them with evidence \(E_k\) into a repair context
\((C_k^{\mathrm{viol}},V_k^{\mathrm{err}},E_k,F_k)\),
and produces an evidence-grounded repair hint that identifies what is inconsistent and why, without directly solving the task (Appendix~\ref{app:hints}).

StateGuard applies progressive hierarchical repair intervention. Light repair provides the repair hint to the agent, while heavy repair additionally removes identified invalid artifacts through harness execution before retrying. The repaired trajectory re-enters the \emph{REPAIR → UPDATE → VERIFY} loop. If repeated intervention still fails, the harness can roll back to a previously valid state or abandon the current state, preventing unresolved errors from contaminating downstream analysis.

% \subsection{Counterfactual State Supervision}
\subsection{Manager-Oriented Counterfactual Supervision}

Existing data-science trajectories are mainly designed for end-to-end task solving~\citep{zhang2025deepanalyzeagenticlargelanguage} and provide limited supervision for analytical-state maintenance, verification, and error repair. As illustrated in Figure \ref{fig:2}(b), we therefore construct manager-oriented trajectories via a counterfactual synthesis pipeline.

\paragraph{Relation-Graph-Based Task Synthesis.} Instead of directly extending existing data-analysis questions into long-horizon tasks, we first construct the underlying state relations, enabling explicit control over analytical dependencies and directly yielding relation annotations for supervision. Starting from raw data-analysis tasks \(\mathcal D_{\mathrm{raw}}\) in DSBench~\citep{jing2025dsbenchfardatascience}, BIRD-INTERACT~\citep{huo2026birdinteractreimaginingtexttosqlevaluation}, and IDA-Bench~\citep{li2025idabenchevaluatingllmsinteractive}, we synthesize a latent state-relation graph \(G^{*}\) and instantiate it into a long-horizon task:
\begin{equation}
\label{eq:task_synthesis}
\mathcal{D}_{\mathrm{raw}}
\xrightarrow{f_{\mathrm{relation}}}
G^{*}
\xrightarrow{f_{\mathrm{task}}}
q^{\mathrm{LH}}.
\end{equation}
The resulting tasks therefore cover \emph{progression}, \emph{branching}, \emph{invalidation}, and \emph{composition} across analytical stages, providing structured supervision for state externalization and relation modeling. 

\paragraph{Executable Counterfactual Runtime Trajectory Synthesis.} Since clean trajectories rarely contain sufficient verification and repair behaviors, we introduce controlled counterfactual interventions into analytical states and replay them with a strong teacher model in an executable runtime:
\begin{equation}
\label{eq:counterfactual_triplet}
\tau^{+}
\xrightarrow{\mathrm{Intervention}}
\tau^{-}
\xrightarrow{\mathrm{Repair}}
\tau^{r},
\end{equation}
where \(\tau^{+}\), \(\tau^{-}\), and \(\tau^{r}\) denote the clean, counterfactual, and repair trajectories. Runtime execution verifies whether an intervention causes a genuine analytical inconsistency and whether repair restores the affected analysis. We retain execution-verified \((\tau^{+},\tau^{-},\tau^{r})\) triplets, while benign interventions serve as hard negatives for conservative decisions. After filtering and quality control, we obtain 3K manager-oriented trajectories that provide structured supervision for StateGuard. Appendix~\ref{app:task_construction} and ~\ref{app:trajectory_generation} provide more details.

\paragraph{Cold-Start via Manager-Oriented SFT.}
Base models cannot reliably follow the StateGuard action protocol or maintain analytical states from sparse reinforcement signals alone. We therefore initialize StateGuard with the synthesized manager-oriented trajectories, mixing \emph{atomic samples} for individual runtime decisions such as state update, verification, and repair with \emph{full trajectories} that preserve complete state lifecycles and dependencies. This enables the model to acquire the structured state protocol and compose individual runtime primitives over long-horizon execution.

\begin{table}[t]
\centering
\caption{Main results on three data-analysis benchmarks. LongDS reports six-domain performance and average accuracy; DAComp-DE covers data implementation and evolution; DABstep reports hard and easy accuracy. DCR measures excess downstream error associated with incorrect upstream. \textbf{Bold} indicates the best result within each group, and --- denotes undefined results.}
\label{tab:1}

\scriptsize
\setlength{\tabcolsep}{1.7pt}
\renewcommand{\arraystretch}{1.12}

\definecolor{groupgray}{RGB}{244,245,247}
\definecolor{sgblue}{RGB}{238,246,252}
\definecolor{pairline}{RGB}{228,230,233}

\resizebox{\linewidth}{!}{
\begin{tabular}{lcccccccccccccc}
\toprule

\multirow[c]{3}{*}{\textbf{Method}}
& \multicolumn{8}{c}{\textbf{LongDS}}
& \multicolumn{4}{c}{\textbf{DAComp-DE}}
& \multicolumn{2}{c}{\textbf{DABstep}} \\

\cmidrule(lr){2-9}
\cmidrule(lr){10-13}
\cmidrule(lr){14-15}

& \multirow[c]{2}{*}{Edu.}
& \multirow[c]{2}{*}{Comm.}
& \multirow[c]{2}{*}{Soc.}
& \multirow[c]{2}{*}{Bus.}
& \multirow[c]{2}{*}{Geo.}
& \multirow[c]{2}{*}{Spo.}
& \multirow[c]{2}{*}{\textbf{Avg.}}
& \multirow[c]{2}{*}{\textbf{DCR}}
& \multicolumn{2}{c}{\textbf{Impl.}}
& \multicolumn{1}{c}{\textbf{Evol.}}
& \multirow[c]{2}{*}{\textbf{DCR}}
& \multirow[c]{2}{*}{Hard}
& \multirow[c]{2}{*}{Easy} \\

\cmidrule(lr){10-12}

& & & & & & & & &
CFS
& CS
& CFS
& &
& \\

\midrule

% =========================================================
% API-based General Agents
% =========================================================

\rowcolor{groupgray}
\multicolumn{15}{l}{\textbf{\textit{Proprietary General Agents}}} \\

GPT-5.5
& 63.56 & 60.30 & 33.76 & 38.60 & 45.40 & 41.50 & 47.14 & 55.34
& 24.68 & 62.14 & 16.10 & 53.10 & 36.77 & 59.72 \\

\rowcolor{sgblue}
\hspace{0.8em}\textbf{+ StateGuard}
& 67.74 & 65.38 & 40.57 & \textbf{42.79} & \textbf{56.84} & \textbf{45.62} & \textbf{54.12} & \textbf{51.25}
& 28.91 & \textbf{67.32} & 21.05 & \textbf{49.96} & 43.65 & 61.11 \\

\arrayrulecolor{pairline}\specialrule{0.5pt}{0pt}{0pt}\arrayrulecolor{black}

Claude Sonnet 5
& 64.52 & 57.48 & 37.13 & 32.54 & 40.29 & 38.76 & 44.33 & 60.56
& 27.64 & 55.37 & 19.41 & 56.20 & 32.28 & 77.78 \\

\rowcolor{sgblue}
\hspace{0.8em}\textbf{+ StateGuard}
& 73.14 & 69.20 & \textbf{42.15} & 36.30 & 52.48 & 45.43 & 53.16 & 56.04
& \textbf{34.51} & 62.56 & \textbf{25.70} & 50.85 & \textbf{35.71} & 79.17 \\

\arrayrulecolor{pairline}\specialrule{0.25pt}{0pt}{0pt}\arrayrulecolor{black}

Kimi-K2.6
& 68.98 & 58.74 & 30.06 & 11.44 & 27.97 & 22.94 & 35.91 & 58.10
& 21.95 & 48.79 & 17.45 & 58.00 & 23.02 & 75.01 \\

\rowcolor{sgblue}
\hspace{0.8em}\textbf{+ StateGuard}
& 71.67 & 62.59 & 36.17 & 28.17 & 35.05 & 27.60 & 44.36 & 55.47
& 25.66 & 56.94 & 21.13 & 56.50 & 32.01 & 80.56 \\

\arrayrulecolor{pairline}\specialrule{0.25pt}{0pt}{0pt}\arrayrulecolor{black}

DeepSeek-V4-Pro
& 70.69 & 64.53 & 38.48 & 14.60 & 43.43 & 43.87 & 44.53 & 62.60
& 30.72 & 57.92 & 14.72 & 57.90 & 25.66 & 81.94 \\

\rowcolor{sgblue}
\hspace{0.8em}\textbf{+ StateGuard}
& \textbf{75.50} & \textbf{73.30} & 39.34 & 33.60 & 55.30 & 42.26 & 54.05 & 56.70
& 32.78 & 64.32 & 17.31 & 55.10 & 30.42 & \textbf{84.72} \\

\midrule

% =========================================================
% Open-Weight General Agents
% =========================================================

\rowcolor{groupgray}
\multicolumn{15}{l}{\textbf{\textit{Open-Weight General Agents}}} \\

Qwen3-30B-A3B
& 19.44 & 3.10 & 6.52 & 0.73 & 4.57 & 0.00 & 5.11 & 41.90
& 0.00 & 4.32 & 11.21 & --- & 3.71 & 70.83 \\

\rowcolor{sgblue}
\hspace{0.8em}\textbf{+ StateGuard}
& \textbf{23.25} & \textbf{5.24} & \textbf{9.94} & 1.85 & \textbf{6.76} & \textbf{4.27} & \textbf{7.66} & 38.50
& \textbf{1.53} & \textbf{7.63} & \textbf{13.92} & --- & 4.23 & \textbf{72.22} \\

\arrayrulecolor{pairline}\specialrule{0.25pt}{0pt}{0pt}\arrayrulecolor{black}

Qwen3-32B
& 20.83 & 3.16 & 6.55 & 1.71 & 3.54 & 0.92 & 5.12 & 34.40
& 0.00 & 0.00 & 12.35 & --- & 2.12 & 36.11 \\

\rowcolor{sgblue}
\hspace{0.8em}\textbf{+ StateGuard}
& 21.09 & 4.27 & 6.42 & \textbf{2.01} & 3.85 & 1.14 & 5.63 & \textbf{33.80}
& --- & --- & 12.56 & --- & 3.44 & 37.50 \\

\arrayrulecolor{pairline}\specialrule{0.25pt}{0pt}{0pt}\arrayrulecolor{black}

Qwen3-8B
& 18.98 & 1.89 & 4.46 & 0.00 & 1.77 & 0.00 & 3.46 & 52.60
& 0.00 & 0.00 & 11.82 & --- & 4.23 & 47.22 \\

\rowcolor{sgblue}
\hspace{0.8em}\textbf{+ StateGuard}
& 16.94 & 3.96 & 7.27 & 0.24 & 3.10 & 0.00 & 4.62 & 52.53
& --- & --- & 12.24 & --- & \textbf{6.08} & 44.45 \\

\midrule

\rowcolor{groupgray}
\multicolumn{15}{l}{\textbf{\textit{Specialized Data Agents}}} \\

DataMind
& 10.65 & 2.11 & 1.49 & 0.24 & 2.36 & 0.00 & 2.47 & 28.00
& 0.00 & 0.00 & 14.56 & --- & 9.79 & 54.17 \\

\rowcolor{sgblue}
\hspace{0.8em}\textbf{+ StateGuard}
& 14.68 & 4.21 & 4.40 & 3.24 & 3.92 & 1.74 & 4.87 & 25.13
& --- & --- & 17.63 & --- & 11.64 & 54.17 \\

\bottomrule
\end{tabular}
}
\end{table}

\subsection{Validity-Guided Policy Optimization}
% \subsection{State Validity Policy Optimization}
While manager-oriented SFT provides a structured cold start, final task outcomes remain insufficient for further optimizing StateGuard: failures may arise from the limited underlying agent even when states are correctly maintained, while successful tasks may still contain unsupported updates or unnecessary interventions. We therefore turn runtime validity evidence into fine-grained signals over protocol correctness, state grounding, and intervention quality.

\paragraph{Validity Runtime Reward.}
For a StateGuard rollout \(\tau\), let the generated tool and control action sequences be \(\mathcal A_{\mathrm{tool}}^\tau=\{a_t^{\mathrm{tool}}\}_{t=1}^{T_{\mathrm{tool}}}\) and \(\mathcal A_{\mathrm{ctrl}}^\tau=\{a_t^{\mathrm{ctrl}}\}_{t=1}^{T_{\mathrm{ctrl}}}\), respectively, and let \(\mathcal S_\tau=\{S_k\}_{k=1}^{K}\) denote the analytical states produced during the rollout. We supervise these objects at three complementary granularities:
\begin{equation}
\label{eq:validity_reward}
\begin{gathered}
R(\tau)
=
\underbrace{\lambda_aR_{\mathrm{action}}+\lambda_lR_{\mathrm{lifecycle}}}_{\text{protocol correctness}}
+\underbrace{\lambda_sR_{\mathrm{support}}}_{\text{state grounding}}
+\underbrace{\lambda_iR_{\mathrm{intervention}}}_{\text{repair quality}},
\\[3pt]
R_{\mathrm{action}}
=\frac{1}{T_{\mathrm{tool}}}\sum_{t=1}^{T_{\mathrm{tool}}}
\mathbb{I}[\mathrm{Exec}(a_t^{\mathrm{tool}})=\mathrm{success}],
\qquad
R_{\mathrm{lifecycle}}
=\frac{1}{T_{\mathrm{ctrl}}}\sum_{t=1}^{T_{\mathrm{ctrl}}}
\mathbb{I}[\mathrm{Valid}(a_t^{\mathrm{ctrl}}\mid\ell_t)],
\\[3pt]
R_{\mathrm{support}}
=
\frac{\sum_{k=1}^{K}\sum_{u\in\mathcal U_k}
\mathbb{I}[\mathrm{Grounded}(u,E_k)]}
{\sum_{k=1}^{K}|\mathcal U_k|}.
\end{gathered}
\end{equation}
The action-level terms reward valid interaction with the StateGuard runtime, where \(\mathrm{Exec}(\cdot)\) checks successful tool execution, \(\mathrm{Valid}(\cdot\mid\ell_t)\) checks whether a control action is legal at lifecycle stage \(\ell_t\).

At the state level, let \(\mathcal{U}_k = V_k \cup N_k\) denote the variables and conclusions recorded in \(S_k\). We reward states whose contents are supported by runtime evidence, and \(\mathrm{Grounded}(u,E_k)\) indicates that \(u\) is supported by tool outputs, executable constraint checks, or the agent's actual execution results.

Finally, the intervention term evaluates whether StateGuard repairs only when warranted.
Let \(g_i\in\{0,1\}\) indicate whether an inconsistency exists at decision point \(i\),
\(d_i\) denote the corresponding control decision, and \(q_i\) measure the
evidence-grounded quality of a triggered repair. We define
\begin{equation}
\label{eq:intervention_reward}
R_{\mathrm{intervention}}
=
\frac{1}{M}\sum_{i=1}^{M}
\left(
R_i^{+}-R_i^{-}
\right),
\end{equation}
where
\(R_i^{+}=q_i\,\mathbb{I}[g_i=1,d_i=\textsc{Repair}]
+\mathbb{I}[g_i=0,d_i\in\{\textsc{Resume},\textsc{Abstain}\}]\)
and
\(R_i^{-}=\beta_{\mathrm{fp}}\mathbb{I}[g_i=0,d_i=\textsc{Repair}]
+\beta_{\mathrm{fn}}\mathbb{I}[g_i=1,d_i\neq\textsc{Repair}]\). 

Here, \(q_i\) evaluates whether the repair is supported by localized evidence and leads to an appropriate correction. We set \(\beta_{\mathrm{fp}}>\beta_{\mathrm{fn}}\) to penalize false repair more heavily than missed intervention, encouraging conservative error correction. Coefficients and evaluation details are given in Appendix~\ref{app:training}.

We optimize StateGuard with DAPO~\citep{yu2025dapoopensourcellmreinforcement} using \(N=8\) rollouts per instance and \(R(\tau)\) as trajectory-level return. Let \(\hat A^{(n)}\) denote the group-normalized advantage, the policy objective is
\begin{equation}
\label{eq:dapo_objective}
\mathcal{J}(\theta)
=
\mathbb E\!\left[
\frac{1}{N}\sum_{n=1}^{N}\sum_t
\min\!\left(
\frac{\pi_\theta}
{\pi_{\theta_{\mathrm{old}}}}
\hat A^{(n)},
\operatorname{clip}\!\left(
\frac{\pi_\theta}
{\pi_{\theta_{\mathrm{old}}}},
1-\epsilon_l,
1+\epsilon_h
\right)
\hat A^{(n)}
\right)
\right].
\end{equation}

\section{Experiments}

\subsection{Experimental Setup}

\paragraph{Benchmarks and Metrics.}
We evaluate StateGuard on three long-horizon data-analysis benchmarks. \textbf{LongDS-Bench}~\citep{xu2026longdsbenchfailurelonghorizonagentic} contains 68 multi-turn tasks and 2,225 queries across six domains with explicit cross-turn dependencies; we report turn-level accuracy. \textbf{DAComp-DE}~\citep{lei2025dacompbenchmarkingdataagents} evaluates repository-level data engineering on implementation and evolution tasks spanning over 4,000 lines of code and 30+ files, where component failures may cascade downstream; we report Component Score (CS) and Cascading Failure Score (CFS) for implementation, and CFS for evolution. \textbf{DABstep}~\citep{egg2025dabstepdataagentbenchmark} contains 450 multi-step tasks (72 easy, 378 hard) requiring code-based processing and contextual reasoning over heterogeneous documentation; we report accuracy on both subsets. We additionally report \textbf{Dependency Contamination Rate (DCR)} to quantify downstream error propagation from incorrect upstream dependencies; lower is better.

\paragraph{Baselines.}
We compare against three families of agents. \textbf{Proprietary general agents} use strong proprietary LLMs, including GPT-5.5, Claude Sonnet 5, Kimi-K2.6, and DeepSeek-V4-Pro. \textbf{Open-Weight general agents} use Qwen3 variants~\citep{qwen3}. All general agents are evaluated under a unified ReAct-style execution interface~\citep{yao2023reactsynergizingreasoningacting} for controlled StateGuard integration. \textbf{Specialized data agents} use DataMind~\citep{qiao2026scalinggeneralistdataanalyticagents}, evaluated under native frameworks. We focus on sufficiently capable models to avoid conflating analytical state management limitations with basic task-execution failure. See details in Appendix~\ref{app:baseline_details}.

\paragraph{Implementation Details.}
We implement StateGuard with Qwen3-8B~\citep{qwen3}, using up to 8 actions per activation, 3 repair attempts per state, and a 40,960-token context. For SFT,  we employ DeepSeek-V4-Pro~\citep{deepseekai2026deepseekv4highlyefficientmilliontoken} to generate 3K trajectories, followed by 3 epochs of training at $5\times10^{-5}$ with batch size 32. RL uses DAPO with 8 rollouts per prompt at $5\times10^{-7}$ and batch size 64. All training uses VeRL~\citep{10.1145/3689031.3696075} on 8 NVIDIA A100 GPUs.

\begin{figure}[t]
\centering

\makebox[\linewidth][c]{%
% =========================================================
% Left: Ablation Table
% =========================================================
\begin{minipage}[c][0.235\textheight][c]{0.60\linewidth}
\centering

\vspace*{0.015\textheight}

\scriptsize
\setlength{\tabcolsep}{2.5pt}
\renewcommand{\arraystretch}{1.08}

\resizebox{\linewidth}{!}{%
\begin{tabular}{lcccccccc}
\toprule

\multirow[c]{3}{*}{
\begin{tabular}[c]{@{}c@{}}
\textbf{Variant}\\[-1pt]
\tiny\textit{Agent: DeepSeek-V4-Pro}
\end{tabular}}
& \multicolumn{2}{c}{\textbf{LongDS}}
& \multicolumn{4}{c}{\textbf{DAComp-DE}}
& \multicolumn{2}{c}{\textbf{DABstep}} \\[-1pt]

\cmidrule(lr){2-3}
\cmidrule(lr){4-7}
\cmidrule(lr){8-9}

& \multirow[c]{2}{*}{\tiny Avg.}
& \multirow[c]{2}{*}{\tiny DCR}
& \multicolumn{2}{c}{\tiny Impl.}
& \multicolumn{1}{c}{\tiny Evol.}
& \multirow[c]{2}{*}{\tiny DCR}
& \multirow[c]{2}{*}{\tiny Hard}
& \multirow[c]{2}{*}{\tiny Easy}
\\[-1pt]

\cmidrule(lr){4-6}

& & &
\tiny CFS &
\tiny CS &
\tiny CFS &
& &
\\

\midrule

\rowcolor{sgblue}
\textbf{StateGuard-8B}
& \textbf{54.05} & \textbf{56.70}
& \textbf{32.78} & \textbf{64.32}
& \textbf{17.31} & \textbf{55.10}
& \textbf{30.42} & \textbf{84.72} \\

\midrule

\rowcolor{groupgray}
\multicolumn{9}{l}{\textbf{\textit{Training}}} \\

Prompt-based
& 47.50 & 61.54 & 30.26 & 58.52
& 14.57 & 57.48 & 25.92 & 83.33 \\

w/o SFT
& 31.65 & 72.91 & 21.69 & 50.90
& 13.11 & 64.20 & 14.02 & 73.61 \\

w/o RL
& 51.78 & 59.94 & 32.10 & 60.12
& 15.68 & 57.19 & 24.88 & 79.16 \\

\midrule

\rowcolor{groupgray}
\multicolumn{9}{l}{\textbf{\textit{Analytical State}}} \\

ReAct baseline
& \emph{44.53} & \emph{62.60}
& \emph{30.72} & \emph{57.92}
& \emph{14.72} & \emph{57.90}
& \emph{25.66} & \emph{81.94} \\

Free-form Memory
& 43.16 & 63.79 & 30.05 & 56.50
& 14.33 & 59.20 & 25.66 & 80.56 \\

w/o State Validity Maintenance
& 40.77 & 68.45 & 24.66 & 51.80
& 13.15 & 60.29 & 21.16 & 76.39 \\

\bottomrule
\end{tabular}%
}

\end{minipage}%
\hspace{0.015\linewidth}%
% =========================================================
% Right: Two Analysis Plots
% =========================================================
\begin{minipage}[c][0.235\textheight][c]{0.375\linewidth}
\centering

\includegraphics[
    width=0.94\linewidth
]{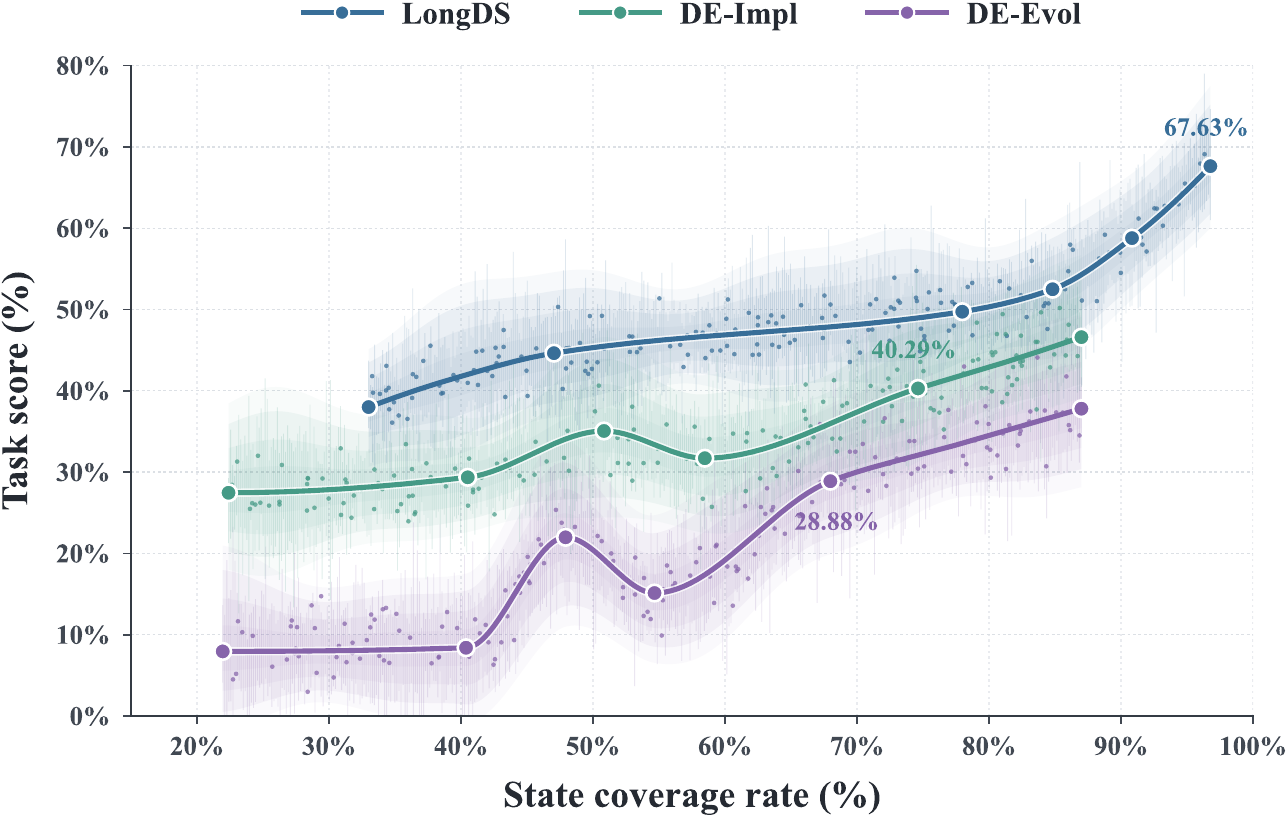}

\vspace{0.60em}

\includegraphics[
    width=0.94\linewidth
]{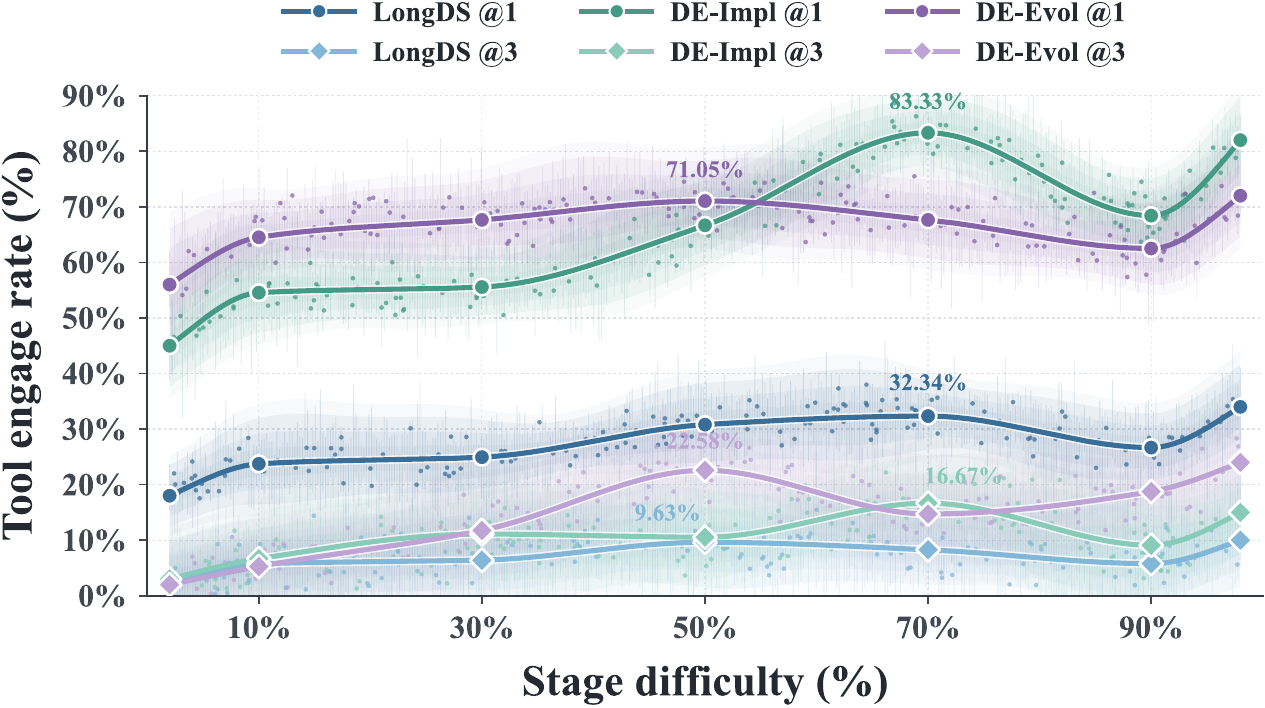}

\end{minipage}%
}

\vspace{0.15em}

\caption{
\textbf{Left:} ablations on training strategy and analytical-state management.
\textbf{Right:} Top: task performance v.s. analytical-state coverage. Bottom: verification engagement across stage difficulty.
}
\label{fig:3}

\end{figure}

\subsection{Main Results}

% Table~\ref{tab:1} shows

\textbf{(1) StateGuard consistently improves long-horizon data analysis across heterogeneous settings.} As shown in Table \ref{tab:1}, StateGuard yields consistent gains over capable backbones in multi-turn analysis, data repository engineering, and multi-step reasoning; for example, it improves DeepSeek-V4-Pro on LongDS (44.53 v.s. 54.05) while reducing DCR. \textbf{(2) Analytical state management complements rather than substitutes for agent capability.} Gains are larger for stronger backbones, while weaker agents benefit less, as StateGuard preserves and verifies analytical progress but does not replace the agent’s basic reasoning and execution capability. \textbf{(3) Broader state maintenance is associated with stronger task performance, while validity verification focuses on more challenging stages.}
As shown in Figure \ref{fig:3} (right), task performance generally increases with analytical-state coverage, while tool verification is more concentrated in medium or high difficulty stages ($\geq 30\%$), especially for intensive process. This suggests StateGuard combines broader state maintenance with selective validity verification to preserve valid artifacts and reduce inconsistencies.

\subsection{Cascading Error Analysis}

To quantify dependency-induced error propagation, for each analytical unit $u_i$, we define $e_i=\mathbb{I}[u_i\text{ is incorrect}]$, $c_i=\mathbb{I}[\exists u_j\in\mathrm{Pa}(u_i),e_j=1]$, $\mathcal{U}_b=\{u_i:c_i=b\}$, and ${\mathrm{DCR}}=\frac{1}{|\mathcal{U}_1|}\sum_{u_i\in\mathcal{U}_1}e_i-\frac{1}{|\mathcal{U}_0|}\sum_{u_i\in\mathcal{U}_0}e_i$, which measures the excess downstream error associated with erroneous upstream. 

\textbf{(1) StateGuard reduces dependency contamination.} On capable backbones, StateGuard consistently lowers DCR; for example, on DeepSeek-V4-Pro, average DCR decreases from 60.25 to 55.90, indicating less downstream inconsistencies associated with state validity and targeted intervention. Trajectory analysis further identifies explicit StateGuard-mediated instances (Appendix~\ref{app:case_analysis}). \textbf{(2) DCR is most informative when the agent has sufficient competence.} For weaker backbones, $|\mathcal{U}_0|$ can approach zero, making DCR low or undefined because limited context windows and local task-solving failures dominate errors rather than lacking of validity. This occurs for Qwen3 variants and DataMind on DAComp-DE, where the repository-level tasks are sufficiently long and difficult that models cannot effectively process the context or solve the base cases correctly. (Appendix~\ref{app:dcr_details}).

\subsection{Ablation Studies}
Figure \ref{fig:3} (left) ablates both training and analytical-state strategies. Training variants remove either SFT or RL, or use prompt-only StateGuard. For state strategies, we compare ReAct, free-form memory, and a variant without externalized states that only provides trajectory-based hints.

\textbf{(1) Training Strategy.}
Prompt-based StateGuard reaches 47.50 on LongDS versus 54.05 for the full model. Removing RL retains partial gains (51.78), whereas removing SFT causes a much larger drop to 31.65. This shows that SFT is essential for acquiring the structured state lifecycle, while validity-guided RL further stabilizes verification and intervention; RL alone is insufficient and may disrupt normal execution. \textbf{(2) Analytical State Validity Management.} Free-form Memory performs similarly to ReAct (43.16 v.s. 44.53), indicating that an additional manager with free memory and checking contributes little. Removing state validity maintenance further drops performance to 40.77 and increases DCR to 68.45, even worse than ReAct. Without persistent states and verifications, hints are unreliable and interfering. The gap to StateGuard confirms explicit state maintenance and validity verification, rather than additional model capacity or generic memory, drive the gains. \textbf{(3) State validity and training provide complementary gains.} Under the same model, performance increases from 40.77 without validity maintenance to 43.16 with free-form memory and 47.50 with structured prompt. SFT and RL further improve it to 51.78 and 54.05, respectively (Appendix~\ref{app:ablation_configurations}).

\section{Case Analysis}
Figure \ref{fig:1} shows a Macrohard equity-IRR case where earlier results are derived under original COGS and a \emph{rounding-to-1K} constraint. \textbf{Linear trajectory.} After the task revises COGS to $0.62$, the agent continues reusing COGS-dependent results, propagating errors to the
final IRR. \textbf{StateGuard} instead externalizes revision as a new state and links it to prior COGS-state through an \textsc{Invalidate} relation, identifies affected artifacts via validity checking and hints targeted recomputation. This illustrates how explicit state relations and targeted intervention prevent stale artifacts from propagating.

\section{Conclusion}
We propose StateGuard, an analytical state validity management framework for long-horizon data agents. Targeting the failure of implicit analytical states and artifacts under the evolving process, StateGuard externalizes analytical progress into explicit states and maintains their validity through evidence verification and hierarchical intervention. Trained with manager-oriented counterfactual supervision and validity-guided policy optimization, StateGuard consistently improves data-agent performance while reducing downstream error propagation across benchmarks. These results demonstrate the value of explicit analytical-state management for long-horizon data analysis.

\subsection*{AI use statement}

Generative AI tools were used to assist with language editing and manuscript refinement, software code development and debugging, and the generation of synthetic training trajectories used in our data construction pipeline. AI-assisted code and generated trajectories were subsequently reviewed, tested, and filtered before being incorporated into the final experiments. No other research components requiring disclosure involved generative AI assistance. The authors reviewed all AI-assisted outputs and take full responsibility for the final content of this work.

\bibliography{iclr2027_conference}
\bibliographystyle{iclr2027_conference}

\appendix
\section{StateGuard Runtime and System Details}

\subsection{Overall StateGuard Runtime Rollout}
\label{app:runtime_rollout}

StateGuard is implemented as an external runtime manager over an underlying
data-analysis agent. Let
\begin{equation}
\label{eq:runtime_state}
\Xi_t =
\bigl(
\tau_{\le t},\,
\mathcal{S}_t,\,
\widetilde{S}_t,\,
\mathcal{R}_t,\,
e_t
\bigr),
\end{equation}
denote the runtime state at step $t$, where $\tau_{\le t}$ is the observed worker
trajectory, $\mathcal{S}_t$ the committed analytical-state store,
$\widetilde{S}_t$ the runtime view of the current draft \(S_k^{\mathrm{draft}}\), $\mathcal{R}_t$ the repair session,
and $e_t$ the current review event. At an adapter-defined review point, the manager
receives
\begin{equation}
\label{eq:manager_observation}
O_t =
\Phi(
\mathcal{T},
\tau_{\le t},
\mathcal{S}_t,
\widetilde{S}_t,
\mathcal{R}_t,
e_t
),
\end{equation}
and produces
\[
a_t^{M} \sim \pi_M(\cdot \mid O_t).
\]
The manager does not directly modify the worker or persistent state. The harness
canonicalizes and validates $a_t^{M}$ and applies the transactional transition
\begin{equation}
\label{eq:harness_transition}
(\Xi_t', o_t)
=
\Gamma_H(\Xi_t, a_t^{M}).
\end{equation}
where
$o_t\in\{\textsc{Continue},\textsc{Finish},\textsc{RunToCompletion}\}$.
Algorithm~\ref{alg:stateguard_rollout} summarizes the runtime procedure.

\begin{algorithm}[h]
\caption{StateGuard runtime rollout.}
\label{alg:stateguard_rollout}
\begin{algorithmic}[1]
\Require Task $\mathcal{T}$; worker policy $\pi_W$; manager policy $\pi_M$;
flow adapter $\mathcal{F}$; harness transition $\Gamma_H$
\Ensure Final answer $y$ and committed analytical states $\mathcal{S}$

\State Initialize worker trajectory $\tau_0 \gets \emptyset$
\State Initialize committed store $\mathcal{S}_0 \gets \emptyset$
\State Initialize draft $\widetilde{S}_0 \gets \varnothing$ and repair session $\mathcal{R}_0$
\State $e_0 \gets \textsc{TaskStart}$, $t \gets 0$

\While{task is not terminated}
    \If{$\mathcal{F}.\textsc{Review}(\tau_{\le t},e_t)=1$}
        \State $O_t \gets
        \Phi(\mathcal{T},\tau_{\le t},\mathcal{S}_t,
        \widetilde{S}_t,\mathcal{R}_t,e_t)$
        \State $a_t^{M}
        \sim \pi_M(\cdot \mid O_t)$
        \Comment{verification tools may be invoked internally}
        \State $(\Xi_t',o_t)
        \gets \Gamma_H(\Xi_t,a_t^{M})$
        \Comment{canonicalize, validate, execute transactionally}

        \If{$o_t=\textsc{Finish}$}
            \State \textbf{break}
        \ElsIf{$o_t=\textsc{RunToCompletion}$}
            \State $y \gets \textsc{RunToCompletion}(\pi_W)$
            \State \textbf{break}
        \Else
            \State $\Xi_t \gets \Xi_t'$
        \EndIf
    \EndIf

    \Repeat
        \State $(r_{t+1},a_{t+1},o_{t+1})
        \sim \pi_W(\cdot \mid \tau_{\le t})$
        \State $\tau_{\le t+1}
        \gets \tau_{\le t}
        \cup \{(r_{t+1},a_{t+1},o_{t+1})\}$
        \State $t \gets t+1$
    \Until{$\mathcal{F}.\textsc{ShouldReview}(\tau_{\le t})=1$
    or worker terminates}

    \State $e_t
    \gets \mathcal{F}.\textsc{ReviewEvent}(\tau_{\le t})$
\EndWhile

\State Finalize runtime trace and discard any uncommitted draft
\State \Return $(y,\mathcal{S})$
\end{algorithmic}
\end{algorithm}

The flow adapter determines the review predicate and event type: turn-based tasks are
reviewed at turn boundaries, whereas fixed-step tasks are reviewed after a configurable
execution window or upon termination. This isolates StateGuard's state-management policy
from benchmark-specific execution schedules.

\subsection{Lifecycle and State Relations}
\label{app:lifecycle_relation}

\paragraph{Lifecycle Adaptation.}
StateGuard supports two lifecycle regimes according to how analytical-state boundaries are
determined. For \emph{turn-aligned} tasks, each user turn corresponds to one analytical state.
Since the query is known before execution, StateGuard opens $S_k$ with its issue,
constraints, and provisional upstream relations before the worker starts. After the worker
completes the turn, StateGuard materializes its variables and conclusions, verifies the state,
and either repairs or commits it:
\[
\textsc{Open}
\rightarrow
\textsc{WorkerTurn}
\rightarrow
\textsc{Update}
\rightarrow
\textsc{Verify}
\rightarrow
\begin{cases}
\textsc{Repair}\rightarrow\textsc{Update},
    & \text{if inconsistent},\\
\textsc{Finalize}\rightarrow\textsc{Commit},
    & \text{otherwise}.
\end{cases}
\]
The complete turn is bound as the source of $S_k$, while provisional relations are finalized
after verification.

For tasks without explicit turn boundaries, StateGuard uses a \emph{segment-induced}
lifecycle. Review points provide observation opportunities but do not themselves define state
boundaries. Let
\[
B_t =
\mathbb{I}
\big[
\text{milestone}
\lor
\text{subgoal completion}
\lor
\text{phase transition}
\big]
\]
denote whether the accumulated trajectory forms a meaningful analytical unit. The lifecycle is
\[
\textsc{Review}
\rightarrow
\begin{cases}
\textsc{Resume}, & B_t=0,\\
\textsc{Open}\rightarrow\textsc{Update}\rightarrow
\textsc{Verify}\rightarrow\textsc{Finalize}\rightarrow\textsc{Commit},
& B_t=1.
\end{cases}
\]
The issue, source interval, and relations are therefore determined after the corresponding
execution segment has been observed.

\paragraph{State Relations.}
For state \(S_k\), let \(\mathcal N_k \subseteq \mathcal S_{<k}\) denote its related committed upstream states:
\[
R_k=\{(S_j,r_{j\to k})\mid S_j\in \mathcal N_k\},\qquad 
r_{j\rightarrow k}\in
\{
\textsc{Init},
\textsc{Progress},
\textsc{Branch},
\textsc{Invalidate},
\textsc{Combine}
\}.
\]

\begin{itemize}[leftmargin=1.5em,itemsep=0pt,topsep=2pt]
    \item \textsc{Init}: no suitable committed upstream state exists.
    \item \textsc{Progress}: directly continues the immediately preceding state.
    \item \textsc{Branch}: starts an alternative path from an earlier, non-adjacent state.
    \item \textsc{Invalidate}: changes an assumption associated with an earlier state, so its
    corresponding result is not inherited by the new branch.
    \item \textsc{Combine}: jointly depends on two or more committed upstream states.
\end{itemize}

For structurally unambiguous cases, the harness normalizes relations automatically. With one
upstream state $S_j$,
\begin{equation}
\label{eq:relation_normalization}
r_{j\rightarrow k} =
\begin{cases}
\textsc{Progress}, & k-j=1,\\
\textsc{Branch},   & k-j>1.
\end{cases}
\end{equation}
whereas multiple upstream states are normalized to \textsc{Combine}. Semantic relations such
as \textsc{Invalidate} remain explicit manager decisions. All upstream references satisfy
\[
\mathcal N_k \subseteq \mathcal S_{<k}
\]
excluding self-references and discarded drafts.

\subsection{Control and Tool Actions}
\label{app:actions}

Following the manager action-space decomposition,
at each manager activation, StateGuard may first acquire targeted verification evidence and
then emits one lifecycle decision:
\begin{equation}
\label{eq:evidence_control}
E_t=\{e_i=\mathcal{T}_i(x_i)\}_{i=1}^{m},
\qquad
a_t^{\mathrm{ctrl}}
\sim
\pi_M(\cdot\mid O_t,E_t),
\end{equation}
where \(T_i\) and \(x_i\) denote the \(i\)-th evidence tool and its input, \(m\) is the number of tool calls at activation \(t\), and \(E_t\) is the activation-local evidence accumulated into the state-level \(E_k\). This instantiates the evidence-guided
verification process in Eq.~\ref{eq:verification_evidence}. Tool actions refine the manager's
evidence; final control action determines lifecycle transition.

\paragraph{Control Actions.}
The lifecycle action space is
\[
\mathcal{A}_{\mathrm{ctrl}}=
\{\textsc{Resume},\textsc{Open},\textsc{Update},\textsc{Finalize},
\textsc{Commit},\textsc{Repair},\textsc{Abandon},\textsc{Abstain}\}.
\]

\begin{sglisting}{StateGuard control-action interfaces}
RESUME_WORKER()

OPEN_STATE(issue, constraints, relations)

UPDATE_STATE(issue, source_interval, used_variables, conclusions)

FINALIZE_RELATIONS(mode, relations, conflict_evidence)

COMMIT_STATE()

REPAIR(analytical_evidence, error_hint)

ABANDON_STATE()

ABSTAIN()
\end{sglisting}

\textsc{OpenState}, \textsc{UpdateState}, \textsc{FinalizeRelations}, and
\textsc{CommitState} construct and persist analytical states;
\textsc{Repair} and \textsc{AbandonState} provide evidence-grounded intervention;
\textsc{ResumeWorker} and \textsc{Abstain} control execution without further state mutation.
The manager only proposes structured decisions. The harness applies
\[
\Xi_{t+1}
=
\Gamma_H(\Xi_t,a_t^{\mathrm{ctrl}}),
\]
enforcing lifecycle legality, dependency validity, and transactional state updates.
Repair requires localized analytical evidence and a structured error hint; repeated repair may
escalate to removal of identified invalid variables from the live workspace before retrying.

\paragraph{Tool Actions.}
StateGuard provides five verification-oriented tools:
\[
\mathcal{A}_{\mathrm{tool}}=
\{\textsc{CheckExecution},\textsc{Inspection},\textsc{Compile},
\textsc{Probe},\textsc{Trace}\}.
\]

\begin{sglisting}{StateGuard verification-tool interfaces}
check_execution(step_id) -> ExecutionResult

inspect_python(code) -> ProgramStructure

compile_python(code) -> SyntaxValidity

run_probe(code) -> ProbeResult

load_state(state_id) -> CommittedState
\end{sglisting}

\texttt{check\_execution} validates whether a worker action received a matching executor
result; \texttt{inspect\_python} extracts program structure through AST analysis;
\texttt{compile\_python} performs compiler-level syntax validation;
\texttt{run\_probe} executes an isolated verification program over task data; and
\texttt{load\_state} retrieves a committed state for direct one-hop dependency checking.
Tools are invoked selectively to resolve concrete validity hypotheses rather than exhaustively
replaying the worker trajectory.

\subsection{Runtime Observation and State Representation}
\label{app:runtime_state}

StateGuard exposes a compact evidence-bearing runtime view rather than the raw worker history.

\paragraph{Runtime Observation.}
At manager activation $t$, the harness constructs
\begin{equation}
\label{eq:runtime_observation}
O_t =
\Big(
e_t,\,
\widehat{\tau}^{\,p}_t,\,
\widetilde{S}_t,\,
\rho_t,\,
r_t^{H},\,
\mathcal{I}_t
\Big),
\end{equation}
where $e_t$ is the lifecycle event,
$\widehat{\tau}^{\,p}_t$ the projected pending trajectory,
$\widetilde{S}_t$ the current draft,
$\rho_t$ the repair and terminal-review status,
$r_t^{H}$ the previous harness result,
and $\mathcal{I}_t$ an optional committed-state index.

\begin{sglisting}{StateGuard runtime observation}
ManagerObservation(
    event_type,
    available_state_id,

    untraced_steps,
    pending_start_step,
    pending_end_step,

    current_draft,
    repair_attempts,
    terminal_pending_review,

    last_action_result,
    committed_state_index
)
\end{sglisting}

The pending trajectory is projected as
\[
\widehat{\tau}^{\,p}_t
=
\Pi_{\mathrm{obs}}
\bigl(
\tau^{p}_t
\bigr),
\]
retaining analytical intent, tool invocation, arguments, and realized execution result while
removing duplicated raw outputs and runtime metadata. Each observed step is therefore represented
as
\[
\hat{\tau}_i
=
\bigl(
i,\,
a_i,\,
o_i
\bigr),
\]
where $a_i$ is the worker action and $o_i$ its executor output or error. The full trajectory is
retained separately for auditing.

The draft $\widetilde{S}_t$ serves as the lifecycle carrier of the current state, while
$r_t^{H}$ provides harness-side feedback for rejected actions or failed executable checks.
Committed history is accessed coarsely through
\begin{equation}
\label{eq:committed_state_index}
\mathcal{I}_t
=
\left\{
\bigl(
\mathrm{id}_j,\,
I_j,\,
N_j
\bigr)
\right\},
\end{equation}
and full state artifacts are retrieved through \texttt{load\_state} only when direct dependency
verification requires additional evidence.

\paragraph{State Representation.}
StateGuard materializes the analytical state defined in
Eq.~\ref{eq:analytical_state}, with additional runtime provenance metadata.
The runtime additionally records provenance metadata such as the supporting worker interval and
checkpoint.

\begin{sglisting}{Analytical-state representation}
AnalyticalState(
    id,
    issue,

    constraints,
    used_variables,
    conclusions,
    relations,

    source_step_start,
    source_step_end,
    checkpoint_id
)

Constraint(
    text,
    code = optional executable check
)

VariableRef(
    name,
    version,
    value
)
\end{sglisting}

Constraints may be executable:
\[
c_i =
\bigl(
\ell_i,\,
\phi_i
\bigr),
\]
where $\ell_i$ is the natural-language specification and $\phi_i$ an optional verification
program. When available, the harness evaluates
\[
\phi_i(D)
\in
\{\textsc{Pass},\textsc{Fail}\}
\]
in an isolated environment over task data, providing directly testable runtime evidence before
commitment.

Variables carry explicit state-level provenance. For $v\in V_k$,
\[
v =
\bigl(
\mathrm{name}(v),\,
\mathrm{ver}(v),\,
\mathrm{value}(v)
\bigr),
\qquad
\mathrm{ver}(v)=S_k,
\]
yielding identifiers such as
\[
\texttt{filtered\_count@S3}.
\]
The version is enforced by the harness, establishing lineage for intermediate analytical
quantities. Drafts remain mutable during construction and repair, whereas committed states are
immutable:
\[
S_k^{\mathrm{draft}}
\xrightarrow{\mathrm{COMMIT}}
S_k.
\]

\subsection{State Hints and Repair Hints}
\label{app:hints}

StateGuard exposes manager-side reasoning to the worker through a restricted guidance interface:
\[
\mathrm{H}
=
\mathrm{H}_{\mathrm{state}}
\cup
\mathrm{H}_{\mathrm{repair}}.
\]
State guidance transfers relevant committed analysis, whereas repair guidance localizes specific
inconsistencies without supplying executable corrections or replacement answers.

\textbf{State Guidance.}
For current state \(S_k\), StateGuard projects the related committed upstream states
\(\mathcal N_k\) into
\begin{equation}
\label{eq:state_guidance}
H_k^{\mathrm{state}} \equiv H_k
=
\mathrm{Pack}(\mathcal N_k),\qquad
\mathrm{Pack}(S_j)
=
\bigl(id_j,I_j,V_j,N_j,R_j\bigr).
\end{equation}
The worker receives the state identifier, issue, variables, conclusions, and relations;
verifier-side constraints, checkpoints, provenance metadata, and source intervals remain hidden.

\begin{sglisting}{Worker-facing analytical-state hint}
<analytical_state_hint>
[
  {
    "id": "S1",
    "issue": "...",
    "variables": [...],
    "conclusions": ["...", "..."],
    "relations": [
      {"type": "progress", "related_state_id": "S0"}
    ]
  }
]
</analytical_state_hint>

These manager-selected states are observations.
For reference only. Verify them before use.
\end{sglisting}

The delivery policy follows the lifecycle. In turn-aligned tasks,
\[
\mathcal N_k=\{S_j\mid(S_j,r_{j\to k})\in R_k\},
\]
so provisional relations directly determine the injected states. In segment-induced tasks,
StateGuard instead refreshes a compact summary of recent committed states after the state store
grows. State guidance therefore provides selective state transfer rather than replaying the full
analytical history.

\paragraph{Repair Guidance.}
When verification identifies a localized inconsistency, StateGuard emits
\begin{equation}
\label{eq:repair_guidance}
H_k^{\mathrm{repair}}
=
\left(
V_k^{\mathrm{err}},
F_k
\right),
\end{equation}
where \(V_k^{\mathrm{err}}\) identifies affected variables or conclusions and
\(F_k\) gives the evidence-grounded faulty-reasoning description.

\begin{sglisting}{Structured repair hint}
<error_hint>
prompt:
  The Manager identified a suspected error in the
  listed variables or conclusions. Review the reason
  below and re-check the affected analysis; this hint
  is only for reference.

error_variable:
  final_answer_ratios

faulty_reasoning:
  The reported ratios use one decimal place although
  the task requires three-decimal precision.
</error_hint>
\end{sglisting}

Repair hints are structurally constrained to remain \emph{non-solution-bearing}: they localize
the affected quantity and faulty reasoning but cannot contain executable repair code or a full
answer. The same error localization also grounds runtime intervention:
\[
\textsc{LightRepair}
=
\textsc{AppendHint},
\]
\[
\textsc{HeavyRepair}
=
\textsc{RemoveInvalidVars}
+
\textsc{AppendHint}.
\]
Repair is append-only with respect to worker context; stronger intervention removes localized
invalid variables from the live workspace before recomputation.

\subsection{Harness and Adapter}

\label{app:harness_adapter}

StateGuard separates semantic decision-making from deterministic runtime enforcement.
The manager proposes analytical content and lifecycle actions, while the harness validates,
canonicalizes, and executes them:
\[
a_t^{M}
\;\xrightarrow{\;\Gamma_H\;}\;
\Xi_{t+1}.
\]
A transition is applied only when its protocol and lifecycle constraints are satisfied,
\begin{equation}
\label{eq:valid_harness_transition}
\Xi_{t+1}
=
\begin{cases}
\Gamma_H(\Xi_t,a_t^{M}),
& \mathrm{Valid}(\Xi_t,a_t^{M})=1,\\
\Xi_t,
& \mathrm{otherwise},
\end{cases}
\end{equation}
where $\mathrm{Valid}(\cdot)$ covers action ordering, terminal conditions, repair budgets,
source intervals, relation references, and flow-specific constraints.

\paragraph{Harnessed Execution.}
The harness owns deterministic bookkeeping that is intentionally excluded from manager
reasoning. It assigns or enforces state identifiers, variable versions, relation normalization,
source binding, repair schedules, interval alignment, and checkpoints. The manager therefore
specifies only analytical semantics, while protocol state remains mechanically controlled.

Each control action is executed as a component-scoped transaction. Let
$\mathcal{C}(a)$ denote the runtime components that action $a$ may modify. The harness applies
\[
\mathrm{snapshot}\!\left(\mathcal{C}(a)\right)
\rightarrow
\Gamma_H(a)
\rightarrow
\begin{cases}
\mathrm{commit}, & \text{on success},\\
\mathrm{restore}\!\left(\mathcal{C}(a)\right), & \text{on failure}.
\end{cases}
\]
The transaction scope may include the worker, workspace, draft state, committed store,
relation graph, and trace buffer, preventing partial runtime mutation without requiring
global rollback.

The harness additionally provides bounded recovery at several levels:
\[
\text{Reject}
\rightarrow
\text{Corrective Feedback}
\rightarrow
\text{Retry},
\]
for invalid manager actions; context overflow triggers progressive observation compression;
transient transport failures receive limited retry; and manager-side failures may degrade gracefully to worker continuation or worker-only
completion, so a manager failure does not necessarily terminate the underlying task.
Repair remains distinct from transactional rollback: Stronger repair instead removes localized invalid variables from the live workspace before recomputation, while repeated repair failure may trigger fallback to a previously valid state.

\paragraph{Adapter Layer.}
Adapters isolate benchmark-specific execution semantics from the StateGuard runtime.
A workflow adapter specifies review timing and lifecycle constraints, while a worker adapter
maps heterogeneous native agents to a common execution contract:
\[
\text{Native Agent}
\;\longrightarrow\;
\{
\texttt{start},
\texttt{step},
\texttt{inject},
\texttt{snapshot},
\texttt{restore},
\texttt{done},
\texttt{final\_answer}
\}.
\]
Benchmark-specific prompting, tool protocols, and termination logic remain inside the adapter;
the harness operates only on the unified worker abstraction. Before execution, StateGuard also
checks that checkpoints, executors, workspaces, and runtime components are consistently bound,
avoiding silent rollback or workspace mismatches.

\subsection{StateGuard Prompting and Runtime Templates}
\label{app:prompt_templates}

StateGuard uses a shared manager controller across benchmarks, with lifecycle-specific
instructions injected by the corresponding flow adapter. The effective manager prompt is
constructed as
\begin{equation}
\label{eq:manager_prompt}
P_M
=
P_{\mathrm{sys}}
\oplus
P_{\mathrm{ctrl}}
\bigl[
P_{\mathrm{life}},
P_{\mathrm{act}}
\bigr]
\oplus
\mathcal{T},
\end{equation}
where $P_{\mathrm{sys}}$ defines the manager role and authority,
$P_{\mathrm{ctrl}}$ specifies the shared StateGuard protocol,
$P_{\mathrm{life}}$ and $P_{\mathrm{act}}$ instantiate flow-specific lifecycle rules,
and $\mathcal{T}$ is the current task. We report below the key prompt fragments that
govern runtime behavior.

\paragraph{Manager System Policy.}
The shared system prompt defines the manager as a state supervisor rather than a task-solving
agent, and explicitly separates evidence acquisition from lifecycle control.

\begin{sglisting}{Manager system prompt: core policy}
You are a State Manager supervising a Worker agent.

1. ROLE
- Observe the Worker trajectory and maintain explicit
  analytical states and relations.
- Detect clear analytical errors, localize their evidence,
  and request limited repair.
- Protect the Worker from unnecessary intervention;
  do not solve the task for it.

2. INFORMATION BOUNDARY
- Use only the task information, Worker trace, execution
  evidence, workspace evidence, current draft, and committed
  states supplied through the runtime.
- Never use or request ground truth, reference answers,
  or judge results.

4. AUTHORITY
- Tool actions obtain missing evidence through FunctionTools.
  They do not end the current Manager activation.
- Control actions make StateGuard lifecycle decisions and
  end the current activation.
- Never call a control action as a tool, and never combine
  a tool action with a control action in one response.
\end{sglisting}

\paragraph{Shared Controller.}
The controller provides the common analytical-state, verification, control, and repair protocol.
Its two flow-dependent slots are
\[
P_{\mathrm{ctrl}}
=
P_{\mathrm{shared}}
\bigl[
\texttt{FLOW\_LIFECYCLE},
\texttt{FLOW\_STATE\_ACTIONS}
\bigr].
\]
The following excerpt captures the shared interaction and verification policy.

\begin{sglisting}{Shared controller: interaction and verification}
Follow the active FLOW-SPECIFIC LIFECYCLE at every review
event. Use only the runtime information supplied to you;
never use ground truth or solve the task for the Worker.
At each activation, take only the next lifecycle action.

Tool actions are read-only FunctionTools. Use only the
focused Tool actions needed, then promptly emit one
control action.

Check only the current state and its relevant Worker trace:

1. Execution: claimed code must have a matching successful
   executor result.
2. Intent and code: repair only a clear mismatch between
   stated intent and code.
3. Code validity: check explicit syntax, execution, or
   missing-variable failures.
4. Query requirements: check clear violations of scope,
   output, format, units, or precision requirements.
5. State consistency: check issue, variables, conclusions,
   the observed Worker span, and directly related-state
   values for concrete conflicts.

Tool actions resolve only specific remaining uncertainty;
they are not a requirement to re-check every step.
\end{sglisting}

The repair policy is similarly evidence-gated: a localized violation with concrete evidence
requires \textsc{Repair}, whereas ambiguous evidence is not sufficient for intervention.
The harness owns the repair schedule and automatically determines the repair attempt and mode.

\begin{sglisting}{Shared controller: repair policy}
Repair a localized violation supported by concrete evidence.

REPAIR and FINALIZE_RELATIONS are the two exits from
checking: a violated constraint you can evidence obliges
REPAIR, and FINALIZE_RELATIONS is correct only when the
checks found no violation.

If evidence is ambiguous, do not REPAIR and do not assert it;
never prescribe a solution, replacement code, ground truth,
or final answer.

The harness owns the repair budget: it chooses light or
heavy, counts attempts automatically, and refuses a repair
once the budget is exhausted.
\end{sglisting}

\paragraph{Flow-Specific Lifecycle Templates.}
Turn-aligned and segment-induced tasks use different lifecycle instructions. For turn-aligned
execution, one turn corresponds to one state and provisional relations are determined before
worker execution.

\begin{sglisting}{Turn-aligned lifecycle: key instructions}
- Every turn is exactly one state.

TURN START:
1. Compare the current query with the compact committed-state
   index and choose provisional relations.
2. OPEN_STATE before Worker execution with the query-defined
   issue, constraints, and provisional relations.
3. After OPEN_STATE succeeds, the harness injects related-state
   hints and starts the Worker.

TURN END:
1. UPDATE_STATE from the completed turn; the harness binds the
   complete turn automatically.
2. Apply the shared checks, then emit REPAIR on clear evidence
   or FINALIZE_RELATIONS otherwise.
3. Confirm provisional relations by default; reselect only on
   explicit conflict.
4. COMMIT_STATE after relation finalization.
\end{sglisting}

For segment-induced execution, review cadence is separated from state formation. The manager
forms a state only when the accumulated trace constitutes a meaningful analytical unit.

\begin{sglisting}{Segment-induced lifecycle: key instructions}
- A pause is not a state boundary.

At each nonterminal pause, decide whether the pending trace
has formed a state. A state has formed when the trace:
- determines an important quantity that later steps rely on;
- completes an explicit sub-stage of the query; or
- transitions from one analytical aspect or sub-goal to another.

If nothing has formed yet, RESUME_WORKER and retain all
pending steps.

If a state has formed:
1. OPEN_STATE.
2. UPDATE_STATE with one inclusive source_interval.
3. Check the written state.
4. Select relations post hoc.
5. REPAIR on clear evidence, otherwise FINALIZE_RELATIONS.
6. COMMIT_STATE.
\end{sglisting}

DAComp uses the same segment-induced lifecycle template, with optional track-specific
verification instructions where required.

\paragraph{Fixed Runtime Guidance.}
Worker-facing state and repair guidance uses fixed reference-only language. The corresponding
runtime strings are:

\begin{sglisting}{Fixed worker-facing guidance}
[Repair hint]
The Manager identified a suspected error in the listed
variables or conclusions. Review the reason below and
re-check the affected analysis; this hint is only for
reference.

[State hint]
These manager-selected states are observations.
For reference only. Verify them before use.
\end{sglisting}

These fixed templates preserve the authority boundary between manager and worker: StateGuard
may expose relevant analytical state or localized error evidence, but does not present manager
outputs as ground truth or replacement solutions. The underlying benchmark workers retain their
native system prompts; StateGuard modifies only the external manager-side runtime and interacts
with the worker through the controlled hint interface.

\section{Training Data and Optimization Details}

\subsection{Source Data}
\label{app:sft_source_data}

\paragraph{Source datasets.}
We construct the SFT source pool from three public data-agent benchmarks:
DSBench~\citep{jing2025dsbenchfardatascience}, BIRD-INTERACT~\citep{huo2026birdinteractreimaginingtexttosqlevaluation},
and IDA-Bench~\citep{li2025idabenchevaluatingllmsinteractive}.
DSBench contains 540 realistic data-science tasks, including 466 data-analysis and
74 data-modeling tasks, spanning long-context, multi-table, and end-to-end modeling
workflows. BIRD-INTERACT provides 600 full and 300 lite interactive database tasks,
covering business-intelligence queries, CRUD operations, dynamic user interaction,
execution feedback, and evolving constraints. IDA-Bench is built from 25 complex
Kaggle notebooks selected from more than 15K candidates and reformulated into
multi-round guided analysis tasks involving iterative data processing, modeling,
and evaluation.

These benchmarks remain challenging for strong contemporary agents and cover
complementary regimes of long-horizon data reasoning: open-ended analysis and
modeling, interactive database operation, and multi-round iterative analysis.
We use their trajectories as the source pool for constructing StateGuard's
state-centric SFT supervision, with diverse long-range dependencies, intermediate
analytical variables, evolving constraints, and recoverable execution errors. The training sources are fully disjoint from the evaluation benchmarks.
\subsection{Task Construction and Counterfactual Intervention Synthesis}
\label{app:task_construction}

Starting from the source tasks in Appendix~\ref{app:sft_source_data}, we instantiate the
relation-graph-based task synthesis summarized in Eq.~\ref{eq:task_synthesis} by constructing
state-centric training instances that expose the latent analytical structure of each task and
augment it with controlled counterfactual branches. The construction follows a
material-first procedure:
\begin{equation}
\label{eq:training_construction}
\mathcal{T}
\;\longrightarrow\;
\{\mathcal{B}^{(1)},\ldots,\mathcal{B}^{(m)}\}
\;\longrightarrow\;
G_{\mathrm{dep}}
\;\longrightarrow\;
\mathcal{S}
\;\longrightarrow\;
\mathcal{S}^{\mathrm{cf}},
\end{equation}
where $\mathcal{B}^{(i)}$ denotes an executable analysis thread,
$G_{\mathrm{dep}}$ its induced dependency graph, $\mathcal{S}$ the resulting analytical-state
trajectory, and $\mathcal{S}^{\mathrm{cf}}$ a counterfactually perturbed trajectory.

\paragraph{Task and Relation-Graph Construction.}
For each task, we first identify multiple executable analytical threads and their material-specific
intermediate results. State dependencies are then reconstructed from actual variable or program
consumption rather than manually assigned relation labels. For an analytical stage $s_i$, let
\[
\mathrm{Dep}(s_i)
=
\{s_j \mid s_i \text{ consumes an output produced by } s_j\}.
\]
The corresponding state relation is determined structurally:
\begin{equation}
\label{eq:relation_construction}
r_i=
\begin{cases}
\textsc{Init}, & |\mathrm{Dep}(s_i)|=0,\\
\textsc{Progress}, &
\mathrm{Dep}(s_i)=\{s_{i-1}\},\\
\textsc{Branch}, &
\mathrm{Dep}(s_i)=\{s_j\},\ j<i-1,\\
\textsc{Combine}, &
|\mathrm{Dep}(s_i)|\ge 2.
\end{cases}
\end{equation}
One concrete realization parses dependencies directly from executable SQL stages by tracing
their \texttt{FROM}/\texttt{JOIN} references. Multiple analytical threads are interleaved before
their results are combined, naturally producing non-local \textsc{Branch} dependencies while
preserving genuine dataflow. The relation-construction rule is aligned with the runtime
normalization used by StateGuard.

\paragraph{Counterfactual Intervention Synthesis.}
We further construct counterfactual branches by perturbing a localized analytical assumption
while preserving the original committed result. Given an analysis
\[
y=f(x;\theta),
\]
we replace a semantically meaningful assumption $\theta$ with an alternative $\theta'$ and
re-execute the affected branch:
\[
y^{\mathrm{cf}}=f(x;\theta').
\]
Interventions cover changes such as numerical parameters, formula semantics, comparison
directions, category scopes, and filtering conditions. They are designed as plausible analytical
alternatives rather than arbitrary corruptions. The resulting state is linked to the original
state through
\[
S_j
\xrightarrow{\;\textsc{Invalidate}\;}
S_k^{\mathrm{cf}},
\]
where the original $S_j$ remains committed and unchanged. This explicitly supervises the
manager to maintain parallel analytical hypotheses instead of overwriting previously valid state.

Together, the original and intervened executions provide clean, counterfactual, and subsequent
repair-oriented trajectories with explicit state contents, dependencies, and evidence, which form
the state-centric supervision used for manager SFT.

\paragraph{Verification and Replay.}
The constructed supervision is validated at three levels. First, executable intermediate results
are materialized and assigned deterministic fingerprints:
\begin{equation}
\label{eq:state_hash}
h_i
=
\mathrm{SHA256}
\left(
\operatorname{sort}
\left\{
\operatorname{canon}(r)
\mid r\in E_i
\right\}
\right),
\end{equation}
where $E_i$ is the result of analytical stage $i$ and
$\operatorname{canon}(\cdot)$ applies deterministic value normalization. The resulting
row-multiset hash is insensitive to row order while remaining sensitive to changes in analytical
content. Second, all agent-visible task files are frozen with SHA-256 hashes after blind
evaluation. Third, the original blind-solving programs are preserved and replayed with path
relocation only; both final answers and saved intermediate calculations must reproduce exactly.
Hidden solutions, reference SQL, state graphs, and evidence artifacts remain outside the
agent-visible task directories.

\subsection{Trajectory Generation and Filtering}
\label{app:trajectory_generation}

To realize the counterfactual trajectory construction summarized in
Eq.~\ref{eq:counterfactual_triplet}, we generate state-centric SFT trajectories using a
\emph{replay-then-live} procedure that preserves a clean analytical prefix while introducing
controlled counterfactual deviations.
Let \(\tau^{+}\) denote a recorded successful trajectory
and \(t^\star\) an intervention point. Generation follows
\[
\tau^{+}
\xrightarrow{\mathrm{replay}}
\tau_{<t^\star}
\xrightarrow{\mathrm{intervene}}
\tau^{-}_{\ge t^\star},
\]
where all worker and manager calls before $t^\star$ must reproduce the parent trajectory
exactly. At $t^\star$, a predefined counterfactual worker action replaces the original action;
subsequent worker and manager behavior is generated live. This isolates the effect of the
intervention from unrelated sampling variation and preserves a well-defined causal lineage
between clean and counterfactual trajectories.

\paragraph{Trajectory Variants.}
From each parent trajectory, we construct several supervision variants depending on which
components remain live after intervention. The principal variants are:
\[
\mathcal{D}_{\mathrm{traj}}
=
\{
\mathcal{D}_{\mathrm{clean}},
\mathcal{D}_{\mathrm{auto}},
\mathcal{D}_{\mathrm{repair}}
\},
\]
where $\mathcal{D}_{\mathrm{clean}}$ contains faithful replays of successful parent trajectories,
$\mathcal{D}_{\mathrm{auto}}$ contains autonomous counterfactual branches in which the manager
must independently detect the injected inconsistency, and $\mathcal{D}_{\mathrm{repair}}$
contains repair demonstrations that supervise part or all of the
\[
\textsc{Detect}
\rightarrow
\textsc{Repair}
\rightarrow
\textsc{Verify}
\rightarrow
\textsc{Commit}
\]
process. Partial demonstrations may constrain only the manager-side repair decision or only the
worker retry, while full demonstrations continue until the corrected analytical state is committed.

\paragraph{Trajectory Filtering.}
We apply both construction-time and runtime validation to remove invalid supervision.
Before execution, injected worker actions must conform to the benchmark's native action protocol,
and forced manager sequences must form valid StateGuard decisions. In particular, a forced
pre-repair manager sequence must terminate with \textsc{Repair}, whereas a forced post-repair
sequence cannot issue another repair and must terminate with \textsc{CommitState}.

After execution, a trajectory is retained only if all required consistency conditions hold:
\begin{equation}
\label{eq:trajectory_filtering}
\mathrm{Keep}(\tau)
=
\mathbb{I}
\left[
\bigwedge_{c\in\mathcal{C}}
c(\tau)=1
\right],
\end{equation}
where $\mathcal{C}$ includes replay fidelity, successful activation of the intended intervention,
complete consumption of any prescribed demonstration, valid lifecycle execution, and task-level
completion. Any mismatch before the intervention point invalidates the branch, since the generated
trajectory would no longer share the intended parent prefix. For repair demonstrations, the
prescribed \textsc{Repair} must additionally be accepted by the harness and induce an actual
worker retry, ensuring that retained examples traverse the same runtime protocol used at inference.

This procedure yields supervision whose clean prefix, intervention, recovery behavior, and final
state transition are all explicitly traceable, while filtering out branches that fail to realize
the intended analytical-state dynamics.

\subsection{Manager-Oriented SFT and Validity-Guided RL}
\label{app:training}

\paragraph{Manager-Oriented SFT.}
We initialize StateGuard with the 3K manager-oriented trajectories constructed above.
The training corpus \(\mathcal D_{\mathrm{SFT}}\) mixes \emph{atomic} samples for individual runtime decisions, such as
state update, verification, relation finalization, and repair, with \emph{complete} trajectories that
preserve full state lifecycles and long-range dependencies. Given a manager observation $x$
and target response $y=(y_1,\ldots,y_L)$, we optimize
\begin{equation}
\label{eq:sft_objective}
\mathcal{L}_{\mathrm{SFT}}
=
-\mathbb{E}_{(x,y)\sim\mathcal{D}_{\mathrm{SFT}}}
\sum_{\ell=1}^{L}
\log
\pi_{\theta}
\left(
y_{\ell}\mid x,y_{<\ell}
\right).
\end{equation}
Atomic supervision teaches the local StateGuard protocol, while full trajectories supervise
the composition of state maintenance, verification, and repair over long-horizon execution.
We train the Qwen3-8B manager for three epochs with a learning rate of
$5\times10^{-5}$ and batch size 32.

\paragraph{Validity-Guided Runtime Reward.}
Task-level outcome alone does not reliably measure manager quality: a task may fail because of
the underlying worker even when StateGuard correctly maintains its analytical states, while a
successful task may still contain unsupported states or unnecessary interventions. We therefore operationalize the validity-guided reward in Eq.~\ref{eq:validity_reward}
using concrete runtime components and coefficients:
\begin{equation}
\label{eq:runtime_reward}
R(\tau)
=
0.15R_{\mathrm{action}}
+
0.20R_{\mathrm{lifecycle}}
+
0.30R_{\mathrm{support}}
+
0.35R_{\mathrm{intervention}}.
\end{equation}

\textbf{Tool-action validity.}
Let $A_{\mathrm{tool}}^\tau
=\{a_t^{\mathrm{tool}}\}_{t=1}^{T_{\mathrm{tool}}}$ be the tool actions generated in
trajectory $\tau$. We measure whether each action is successfully parsed by the harness and
produces a valid tool invocation:
\begin{equation}
\label{eq:tool_reward}
R_{\mathrm{action}}
=
\frac{1}{T_{\mathrm{tool}}}
\sum_{t=1}^{T_{\mathrm{tool}}}
\mathbb I[
\mathrm{Exec}(a_t^{\mathrm{tool}})=\mathrm{success}
].
\end{equation}
This term provides deterministic supervision for the structured tool protocol independently
of downstream task success.

\textbf{Lifecycle correctness.}
For control actions
$A_{\mathrm{ctrl}}^\tau
=\{a_t^{\mathrm{ctrl}}\}_{t=1}^{T_{\mathrm{ctrl}}}$,
let $\ell_t$ denote the active lifecycle stage. We define
\begin{equation}
\label{eq:lifecycle_reward}
R_{\mathrm{lifecycle}}
=
\frac{1}{T_{\mathrm{ctrl}}}
\sum_{t=1}^{T_{\mathrm{ctrl}}}
\mathbb{I}
\left[
\operatorname{Valid}
\left(
a_t^{\mathrm{ctrl}}
\mid
\ell_t
\right)
\right].
\end{equation}
A control action contributes positively only when it respects the StateGuard lifecycle enforced
by the harness.

\textbf{State support.}
For each analytical state $S_k$, let
\[
U_k = V_k \cup N_k
\]
denote its recorded variables and conclusions. We reward only assertions with concrete runtime
support:
\begin{equation}
\label{eq:support_reward}
R_{\mathrm{support}}
=
\frac{
\displaystyle
\sum_{k=1}^{K}
\sum_{u\in U_k}
\mathbb{I}
\left[
\operatorname{Grounded}(u,E_k)
\right]
}{
\displaystyle
\sum_{k=1}^{K}|U_k|
},
\end{equation}
where $E_k$ denotes the available execution and verification evidence. An assertion is considered
supported when the corresponding value or conclusion is grounded by worker execution results,
manager verification-tool outputs, or executable constraint checks. Pure natural-language
reasoning or planning that has not been verified by execution does not count as support, even
if the manager subsequently records it as a state assertion. For cases that cannot be determined
mechanically, support is evaluated by the LLM judge.

\paragraph{Intervention quality.}
The final reward component evaluates whether StateGuard intervenes only
when an actual inconsistency is present and whether the resulting repair is useful.
For a rollout with \(M\) intervention decision points, we instantiate
\(R_{\mathrm{intervention}}\) in Eq.~\ref{eq:intervention_reward} with the following repair-quality score and penalties:
\begin{equation}
q_i=
\begin{cases}
1.0, & \text{correct repair of a genuine inconsistency},\\
0.2, & \text{appropriate but only partially effective repair},\\
-0.5, & \text{misleading or incorrectly localized repair},
\end{cases}
\qquad
\beta_{\mathrm{fp}}=1.0,\quad
\beta_{\mathrm{fn}}=0.3.
\end{equation}
Semantic detection and repair quality are evaluated with GPT-OSS-120B when they cannot be resolved
deterministically from runtime evidence.

\paragraph{Validity-Guided Policy Optimization.}
We initialize reinforcement learning from the SFT checkpoint and optimize the manager with
DAPO~\citep{yu2025dapoopensourcellmreinforcement}. For each training instance, we generate $N=8$ manager rollouts.
To isolate manager quality from stochastic variation in the underlying data-analysis agent,
rollouts within the same DAPO group share the same worker-side prefix and runtime context.
Only the manager continuation is resampled:
\[
\tau_M^{(n)}
\sim
\pi_{\theta}
\left(
\cdot
\mid
x^{W}
\right),
\qquad
n=1,\ldots,N,
\]
where $x^{W}$ contains the fixed worker trajectory prefix together with the corresponding
StateGuard runtime observation. Thus, reward differences within a group primarily reflect
alternative state-management, verification, and repair decisions rather than independent worker
executions.

Let
\[
\widehat{A}^{(n)}
=
\operatorname{Norm}
\left(
R(\tau_M^{(n)});
\{R(\tau_M^{(j)})\}_{j=1}^{N}
\right)
\]
denote the group-normalized advantage and
\[
\rho_t^{(n)}(\theta)
=
\frac{
\pi_{\theta}
\left(
a_t^{(n)}\mid s_t^{(n)}
\right)
}{
\pi_{\theta_{\mathrm{old}}}
\left(
a_t^{(n)}\mid s_t^{(n)}
\right)
}.
\]
Following DAPO, the policy is optimized with the clipped objective (Eq.~\ref{eq:dapo_objective})
\[
\mathcal{J}(\theta)
=
\mathbb{E}
\left[
\frac{1}{N}
\sum_{n=1}^{N}
\sum_t
\min
\left(
\rho_t^{(n)}(\theta)\widehat{A}^{(n)},
\operatorname{clip}
\left(
\rho_t^{(n)}(\theta),
1-\epsilon_l,
1+\epsilon_h
\right)
\widehat{A}^{(n)}
\right)
\right].
\]
where \(\epsilon_l\) and \(\epsilon_h\) denote the lower and upper clipping bounds. This controlled grouped rollout design places optimization pressure directly on StateGuard's
runtime behavior under a common analytical context.

\begin{table}[t]
\centering
\small
\caption{Training configuration and runtime reward settings for StateGuard.}
\label{tab:stateguard_training}
\begin{tabular}{lll}
\toprule
Category & Hyperparameter & Value \\
\midrule
\multirow{4}{*}{SFT}
& Backbone & Qwen3-8B \\
& Training trajectories & 3K \\
& Epochs & 3 \\
& Learning rate & $5\times10^{-5}$ \\
& Batch size & 32 \\
\midrule
\multirow{7}{*}{RL}
& Algorithm & DAPO \\
& Framework & VeRL \\
& Rollouts per prompt & 8 \\
& Learning rate & $5\times10^{-7}$ \\
& Batch size & 64 \\
& Reward judge & GPT-OSS-120B \\
& Hardware & 8$\times$ NVIDIA A100 \\
\midrule
\multirow{4}{*}{Reward}
& Tool-action weight & 0.15 \\
& Lifecycle weight & 0.20 \\
& State-support weight & 0.30 \\
& Intervention-quality weight & 0.35 \\
\midrule
Runtime
& Context window & 40,960 \\
\bottomrule
\end{tabular}
\end{table}

\section{Experimental Protocol}
\subsection{Benchmark Details}
\label{app:benchmark_details}

\paragraph{LongDS-Bench.}
LongDS-Bench~\citep{xu2026longdsbenchfailurelonghorizonagentic} evaluates long-horizon, multi-turn data analysis under evolving analytical state. The benchmark contains 68 tasks and 2,225 turns across six domains: Education, Commerce, Society, Business, Geoscience, and Sports. Its tasks are designed to require persistent reasoning over distant turns, including state updates, counterfactual revisions, rollback, and multi-stage composition. We evaluate on the full benchmark and report turn-level accuracy for each domain together with the overall average. LongDS is particularly suitable for evaluating StateGuard because success depends on preserving and updating previously established analytical artifacts across long interaction horizons.

\paragraph{DAComp-DE.}
DAComp~\citep{lei2025dacompbenchmarkingdataagents} evaluates data agents across realistic data-intelligence workflows. We use its Data Engineering setting, which contains repository-level tasks involving multi-file codebases and long dependency chains. Our experiments focus on the executable Implementation and Evolution tracks. Implementation requires constructing multi-stage data pipelines from task specifications, while Evolution requires modifying existing data systems under changed requirements. Individual tasks may involve more than 4,000 lines of code distributed across over 30 files, making upstream failures likely to propagate across dependent components. Following the benchmark protocol, we report Component Score (CS) and Cascading Failure Score (CFS) for Implementation, and CFS for Evolution.

\paragraph{DABstep.}
DABstep~\citep{egg2025dabstepdataagentbenchmark} contains 450 realistic multi-step data-analysis tasks, including 72 easy and 378 hard instances. The benchmark combines structured data files with heterogeneous contextual documentation and requires code-based processing, multi-source reasoning, and precise answer generation. Easy tasks primarily involve relatively direct lookup, filtering, or aggregation, whereas hard tasks require substantially deeper multi-step computation and contextual reasoning across multiple sources. Evaluation is fully automatic and does not rely on an LLM judge. We report accuracy separately on the easy and hard subsets.

\subsection{Baseline Details}
\label{app:baseline_details}

We compare StateGuard against three families of data agents. \emph{Proprietary general agents}
include GPT-5.5, Claude Sonnet 5, Kimi-K2.6, and DeepSeek-V4-Pro. \emph{Open-weight general
agents} include Qwen3-30B-A3B, Qwen3-32B, and Qwen3-8B. We additionally evaluate the specialized
data agent DataMind under its native framework. For general-purpose models, all experiments use
the same benchmark-specific worker implementation and ReAct-style execution interface as their
corresponding StateGuard runs.

\paragraph{Unified Worker and Harness.}
Baseline and StateGuard runs share the same benchmark adapter, worker implementation, execution
environment, input files, and native worker prompt. The baseline is obtained by disabling the
manager:
\[
\text{Baseline}
=
\text{Worker}+\text{Harness},
\qquad
\text{StateGuard}
=
\text{Worker}+\text{Harness}+\text{Manager}.
\]
When the manager is absent, the harness executes a pure worker trajectory until termination and
does not construct manager observations, maintain analytical states, invoke validity verification,
or inject state/repair hints. StateGuard therefore augments the same underlying data agent through
an external state-management pathway rather than replacing its reasoning or execution policy.

All benchmark workers preserve their official prompting conventions. LongDS directly uses the
official DSGym prompt templates; DABstep follows the prompt construction and reasoning-model
handling of its official runner; and DAComp-DE retains the native OpenHands/CodeActAgent system
prompt. The adapters do not introduce an alternative worker system prompt. StateGuard enters the
worker context only through the runtime state and repair hints described in
Appendix~\ref{app:prompt_templates}.

\paragraph{Interaction Budgets.}
Manager actions are executed outside the worker action budget and are therefore not counted as
worker steps. For settings in which StateGuard may request the worker to inspect or recompute an
affected analysis, we allow a slightly larger worker budget so that the worker has sufficient
execution capacity to respond to manager feedback. The resulting configurations are summarized
below.

\begin{table}[H]
\centering
\small
\caption{Worker interaction budgets used for baseline and StateGuard evaluation. Manager actions
do not consume worker steps.}
\label{tab:baseline_runtime}
\begin{tabular}{lccc}
\toprule
Benchmark & Baseline Worker Budget & +StateGuard Worker Budget & Manager Review \\
\midrule
LongDS-Bench & 40 / turn & 40 / turn & Turn boundary \\
DABstep & 10 & 15 & Every 3 worker steps \\
DAComp-DE & 50 & 60 & Every 5 worker steps \\
\bottomrule
\end{tabular}
\end{table}

The additional worker budget in DABstep and DAComp-DE is reserved only for
manager-triggered re-analysis or correction. In practice, 78.6\% of DABstep runs,
83.3\% of DAComp-DE Implementation runs, and 90.0\% of Evolution runs remain within
the original baseline worker-step limits, indicating that the enlarged caps mainly provide
recovery headroom rather than routine additional computation. LongDS uses identical
worker budgets for baseline and StateGuard.

\subsection{Evaluation Details}
\label{app:evaluation_details}

We follow the official evaluation protocol of each benchmark. Since the three benchmarks differ substantially in their scoring unit and matching criterion, we report their metrics separately rather than treating them as directly interchangeable.

\paragraph{LongDS-Bench Evaluation.}
LongDS-Bench is evaluated with the official DSGym LLM-based judge. We directly invoke the benchmark's native evaluation function and use DeepSeek-V4-Pro as the judge model for all experimental arms. The evaluation unit is an individual turn rather than a complete task. For each turn, the judge compares the model response against the required ground-truth fields and assigns a binary score
\[
s_t \in \{0,1\}.
\]
The final LongDS score is the micro-average over all judged turns,
\[
\mathrm{Acc}_{\mathrm{LongDS}}
=
\frac{1}{T}
\sum_{t=1}^{T}s_t,
\]
where $T=2{,}225$ over the full benchmark.

The official judge evaluates only fields explicitly requested by the current question. Required numerical values must match exactly except for insignificant trailing zeros, unless the ground truth explicitly specifies a numerical tolerance. Ordered outputs are additionally checked for ordering when ranking is part of the task, while superficial differences in capitalization, punctuation, spacing, labels, currency symbols, or explanatory wording are ignored. A turn receives a score of one only when all required fields are correct; otherwise it receives zero. We use the same judge model and official protocol for every baseline and StateGuard configuration.

\paragraph{DABstep Evaluation.}
DABstep uses its official programmatic scorer with withheld ground-truth answers. Evaluation is performed per task and produces a binary correctness value. The benchmark contains 72 easy and 378 hard instances, and we report accuracy separately for the two subsets:
\[
\mathrm{Acc}_{g}
=
\frac{1}{|\mathcal{D}_{g}|}
\sum_{i\in\mathcal{D}_{g}}
\mathbb{I}
\left[
\hat{y}_{i}\simeq y_i
\right],
\qquad
g\in\{\mathrm{easy},\mathrm{hard}\},
\]
where $\simeq$ denotes the benchmark's normalized answer-matching procedure.

The official scorer dispatches answers to numeric, list, or string comparison according to their format. Numeric values below one are compared using
\[
\mathrm{rel\_tol}=\mathrm{abs\_tol}=10^{-4},
\]
while larger values are compared after normalization to the smaller number of displayed decimal places. Lists are normalized and compared without regard to element order. String answers are lower-cased and normalized by removing punctuation and other superficial formatting differences before comparison. We use the official scorer without introducing an additional LLM judge.

\paragraph{DAComp-DE Evaluation.}
DAComp-DE is evaluated programmatically by rebuilding the submitted data-engineering pipeline and comparing the resulting database against the benchmark reference. Evaluation is performed at the table level under a hierarchical weighting scheme:
\[
\text{Layer}
\;\rightarrow\;
\text{Table}
\;\rightarrow\;
\text{Compared Columns}.
\]
Each analytical layer is assigned a benchmark-defined weight, individual tables receive weights within the layer, and correctness is computed only over the configured comparison columns. This produces a continuous task score in the range $[0,100]$ rather than a binary outcome.

Before table-level scoring, each submission must satisfy two execution gates:
\[
\operatorname{RunSuccess}=1,
\qquad
\operatorname{SchemaValid}=1,
\]
ensuring that the submitted pipeline executes successfully and produces the required database schemas. We evaluate Cascading Failure Score (CFS) by rebuilding and executing the complete predicted pipeline from a clean state. Consequently, an incorrect upstream table can propagate naturally into dependent downstream tables, preserving the dependency-induced failure behavior that StateGuard is designed to mitigate. Where reported in the main results, Component Score (CS) evaluates individual components under the benchmark's corresponding isolated protocol.

\paragraph{Metric Semantics.}
The three benchmarks therefore expose complementary evaluation granularities:
\[
\begin{array}{ll}
\text{LongDS} &
\text{turn-level binary LLM judgment},\\
\text{DABstep} &
\text{task-level binary programmatic matching},\\
\text{DAComp-DE} &
\text{weighted execution-based table scoring}.
\end{array}
\]
Accordingly, performance differences should be interpreted within each benchmark's native metric rather than as numerically equivalent changes across benchmarks.

LongDS is particularly sensitive to upstream corrections: because dependent turns are scored
individually, fixing an early analytical error can improve several downstream turns simultaneously,
leading to larger aggregate score shifts under the turn-level micro-average.

\paragraph{Statistical Robustness.}
Because turn outcomes are correlated within tasks, we assess uncertainty using a paired
cluster bootstrap over the 68 LongDS tasks, preserving all turns within each sampled task.
On a matched DeepSeek-V4-Pro run covering all 2,225 turns, StateGuard improves
micro-accuracy by 9.44 points, with a 95\% confidence interval of $[1.69, 14.13]$
over 10,000 bootstrap resamples; only 0.59\% of resamples yield a non-positive gain.
The corresponding task-macro gain is 9.52 points, closely matching the micro estimate
and indicating that the improvement is not driven by a small number of long tasks.

\subsection{DCR Details}
\label{app:dcr_details}

Dependency Contamination Rate (DCR) measures the excess downstream error associated with
incorrect upstream analytical dependencies. Unlike benchmark answer evaluation, DCR itself is
computed entirely programmatically and does not involve an LLM judge or an additional evaluation.

\paragraph{Formal Definition.}
Let $\mathcal{U}$ denote the set of analytical units in a benchmark. The unit is benchmark-specific:
a \emph{turn} in LongDS and a derived \emph{table} in DAComp-DE. For each unit $u\in\mathcal{U}$,
let
\[
e_u
=
\mathbb{I}
[u\text{ is incorrect}]
\]
denote its correctness indicator, and let $\mathrm{Pa}(u)$ denote its declared upstream dependencies.

Because some dependencies may not have an observed correctness value, we first define the
observable parent set
\[
\widetilde{\mathrm{Pa}}(u)
=
\left\{
v\in\mathrm{Pa}(u)
\mid
e_v\text{ is observed}
\right\}.
\]
Only units satisfying
\[
\left|
\widetilde{\mathrm{Pa}}(u)
\right|>0
\]
participate in DCR computation. We then define the contamination indicator
\[
c_u
=
\mathbb{I}
\left[
\exists v\in\widetilde{\mathrm{Pa}}(u)
\text{ such that }e_v=1
\right].
\]
Accordingly,
\[
\mathcal{U}_1
=
\{u\in\mathcal{U}:c_u=1\},
\qquad
\mathcal{U}_0
=
\{u\in\mathcal{U}:c_u=0\},
\]
represent units with at least one incorrect upstream dependency and units whose observed upstream
dependencies are all correct, respectively. DCR is
\begin{equation}
\label{eq:dcr}
\mathrm{DCR}
=
\underbrace{
\frac{1}{|\mathcal{U}_1|}
\sum_{u\in\mathcal{U}_1}e_u
}_{P(e_u=1\mid c_u=1)}
-
\underbrace{
\frac{1}{|\mathcal{U}_0|}
\sum_{u\in\mathcal{U}_0}e_u
}_{P(e_u=1\mid c_u=0)}.
\end{equation}
Thus, DCR directly estimates the additional downstream error probability associated with an
incorrect declared dependency. If either $\mathcal{U}_1$ or $\mathcal{U}_0$ is empty, DCR is
reported as undefined rather than assigning an artificial value.

For auxiliary cross-model analysis, we may additionally inspect the corresponding risk ratio
\[
\mathrm{RR}
=
\frac{
P(e_u=1\mid c_u=1)
}{
P(e_u=1\mid c_u=0)
},
\]
which normalizes contamination by the model's baseline downstream error rate. DCR remains the
primary metric reported in the main experiments.

\paragraph{LongDS: Turn-Level Dependency Contamination.}
For LongDS, each analytical unit corresponds to one evaluated turn. Correctness is inherited from
the official turn-level evaluation:
\[
e_t = 1-s_t,
\qquad
s_t\in\{0,1\}.
\]
Only turns with a successfully produced binary judge score are considered. A missing or failed
evaluation is not interpreted as an incorrect answer. Likewise, an upstream dependency contributes
to the contamination condition only when that dependency has itself been successfully evaluated.

The computation can be summarized as follows:

\begin{sglisting}{LongDS DCR Computation}
scores = {
    turn_id: judge_score
    for turn_id, judge_score in evaluated_turns
    if judge_score in {0, 1}
}

contaminated_bad = contaminated_n = 0
clean_bad = clean_n = 0

for turn, declared_deps in dependency_map.items():
    if turn not in scores:
        continue

    observed_deps = [d for d in declared_deps if d in scores]
    if not observed_deps:
        continue

    wrong = 1 - scores[turn]

    if all(scores[d] == 1 for d in observed_deps):
        clean_n += 1
        clean_bad += wrong
    else:
        contaminated_n += 1
        contaminated_bad += wrong

if contaminated_n == 0 or clean_n == 0:
    DCR = undefined
else:
    p_contaminated = contaminated_bad / contaminated_n
    p_clean = clean_bad / clean_n
    DCR = p_contaminated - p_clean
\end{sglisting}

The remaining problem is to recover $\mathrm{Pa}(t)$ from the benchmark reference artifacts.
LongDS declares cross-turn dependencies in comments within its reference programs, but the
surface form is heterogeneous, including expressions such as
\texttt{Depends on Task 3},
\texttt{Tasks 1, 8, and 9},
\texttt{from Task 5},
and \texttt{Tasks 2--4}.
We therefore construct a deterministic dependency parser over the benchmark reference comments.

Conceptually, dependency extraction is
\[
\mathrm{Pa}(t)
=
\operatorname{ParseRefs}
\left(
\operatorname{Comments}(t)
\right)
\cap
\{1,\ldots,t-1\}.
\]
The parser normalizes singular references, lists, conjunctions, slash-separated references, and
bounded numerical ranges. Self-references and references to future turns are removed, since a
later turn cannot be an upstream dependency of the current turn. A simplified implementation is:

\begin{sglisting}{LongDS Dependency Extraction}
def extract_dependencies(turn_id, comments):
    refs = set()

    for comment in comments:
        comment = remove_current_turn_prefix(comment)

        # Accept heterogeneous forms such as:
        # Task 3
        # Tasks 1, 8, and 9
        # Task 14/17
        # Tasks 2-4
        refs |= parse_task_reference_lists(comment)

    refs.discard(turn_id)

    # Only earlier turns can be upstream dependencies.
    refs = {d for d in refs if 0 < d < turn_id}

    return sorted(refs)
\end{sglisting}

For range expressions, we apply a conservative expansion rule and reject implausibly large spans
to avoid interpreting unrelated numeric text as a dependency interval. The extracted dependency
map is manually inspected once, serialized into a frozen mapping
\[
G_{\mathrm{LongDS}}
=
\{t\mapsto\mathrm{Pa}(t)\},
\]
and reused unchanged for every model and StateGuard arm. This ensures that differences in DCR are
caused by model behavior rather than by model-specific dependency annotation.

\paragraph{DAComp-DE: Table-Level Dependency Contamination.}
For DAComp-DE Implementation, the analytical unit is a derived database table. In contrast to
LongDS, dependency structure is recovered directly from executable reference SQL rather than language annotations. For table $T_i$, we define
\[
\mathrm{Pa}(T_i)
=
\left\{
T_j:
T_j
\text{ is referenced by a }
\texttt{FROM}
\text{ or }
\texttt{JOIN}
\text{ clause of }T_i
\right\}.
\]
Before parsing, SQL line comments are removed. We then retain only references to tables defined
within the same benchmark task and remove self-references. External raw-source tables are therefore
excluded automatically from the analytical dependency graph.

\begin{sglisting}{DAComp-DE Dependency Extraction}
def table_dependencies(task):
    own_tables = all_reference_tables(task)
    deps = {}

    for sql_file in reference_sql_files(task):
        table = table_name(sql_file)
        sql = strip_line_comments(read(sql_file))

        referenced = parse_from_and_join_targets(sql)

        deps[table] = sorted({
            ref
            for ref in referenced
            if ref in own_tables and ref != table
        })

    return deps
\end{sglisting}

Table correctness is taken from the benchmark's CFS evaluation output. Let
\[
e_{T_i}
=
\mathbb{I}
[
T_i\text{ fails its CFS table comparison}
].
\]
The same contaminated-versus-clean partition used for LongDS is then applied at the table level:

\begin{sglisting}{DAComp-DE Table-Level DCR}
for table, declared_deps in dependency_graph.items():
    if table not in table_scores:
        continue

    observed_deps = [
        d for d in declared_deps
        if d in table_scores
    ]
    if not observed_deps:
        continue

    wrong = int(not table_scores[table])

    if all(table_scores[d] for d in observed_deps):
        clean_n += 1
        clean_bad += wrong
    else:
        contaminated_n += 1
        contaminated_bad += wrong
\end{sglisting}

Using CFS correctness is essential for this analysis. CFS evaluates each table inside the
model-produced pipeline, so an incorrect upstream table is precisely the value that reaches its
downstream consumers. In contrast, Component Score evaluates components under isolated reference
conditions and can replace upstream context with gold components. Such evaluation would remove
the realized contamination path and therefore cannot faithfully instantiate
$P(e_u=1\mid c_u=1)$.

\paragraph{Valid Units and Benchmark Scope.}
Units without declared analytical dependencies are excluded from both conditional groups. This
typically includes initial LongDS turns and DAComp-DE staging tables that directly consume raw
sources. Similarly, if a dependency has no observed correctness value, it is omitted from the
observable parent set rather than treated as correct or incorrect.

DCR is not computed for DABstep because its tasks do not expose an explicit inter-unit dependency
structure comparable to LongDS turns or DAComp-DE derived tables. Overall, the DCR pipeline is
fully deterministic:
\[
\text{declared dependencies}
\rightarrow
\text{upstream correctness}
\rightarrow
\text{clean/contaminated partition}
\rightarrow
\text{downstream error gap}.
\]
This isolates dependency-induced error propagation from ordinary local task failure and directly
measures whether incorrect analytical artifacts are more likely to contaminate their downstream
dependents.

\subsection{Ablation Configurations}
\label{app:ablation_configurations}

We conduct ablations along two axes: training strategy and analytical-state design. For training,
we compare the full model against \textsc{w/o RL}, which uses the SFT checkpoint directly, and
prompt-only variants based on the stock Qwen3-8B manager. The main ablations focus on the state
mechanism itself and form a controlled hierarchy from stateless observation to unrestricted memory
and finally structured StateGuard.

\paragraph{w/o State Validity Maintenance.}
This variant removes the entire analytical-state and validity-management mechanism. The manager
has no persistent state, lifecycle, verification tools, executable constraints, repair action, or
cross-activation memory. At each review point, a fresh Qwen3-8B observer receives only the recent
worker-step window and may emit a single non-solution-bearing hint:
\[
h_t
=
f_{\mathrm{obs}}
\left(
\tau_t^{\mathrm{window}}
\right),
\qquad
h_t\in\{\varnothing,\text{hint}\}.
\]
The observer session is reconstructed at every activation, so information cannot persist across
review points. This isolates the effect of adding an auxiliary model that locally monitors the
worker without maintaining analytical state.

\begin{sglisting}{Prompt excerpt for w/o State Validity Maintenance}
You watch a Worker agent solve a data-analysis task. You are shown a window of
its most recent steps and nothing else.

You have no memory: each activation stands alone. You have no tools and cannot
execute anything, so never claim to have verified a step or invent an execution
result.

Reading only the steps in this window, is there anything worth telling the
Worker before it continues?

Reply with one JSON object:
{"hint": null}
or
{"hint": "a localized remark grounded in the observed worker steps"}
\end{sglisting}

Only the local worker trace, remaining worker budget, and termination status are exposed; draft
states, committed-state history, and previous manager outputs are withheld. When no grounded issue
is identified, the worker resumes without intervention.

\paragraph{Free-Form Memory.}
This variant retains the full StateGuard runtime---including the state lifecycle, state schema,
cross-state relations, verification tools, persistent memory, and repair channel---but removes the
prescriptive state-content and verification policies. The manager may populate the existing state
fields in any useful form, leave them empty, attach executable constraint code or not, and invoke
verification tools at arbitrary points or not at all. The harness schema is unchanged, so the
ablation removes semantic guidance rather than interface structure.

\begin{sglisting}{Prompt excerpt for Free-Form Memory}
State content -- yours to decide.

Nothing is prescribed about what belongs in the state fields. There is no
required content and no rule specifying which quantities deserve a constraint,
which variables are worth naming, or how a conclusion should be phrased.

The field shapes accepted by the harness remain fixed:
  constraint: {"text":"..."}
              {"text":"...","code":"<standalone python>"}
  variable:   {"name":"filtered_count","value":117}
  conclusions:["..."]

Checking is also up to you. No checklist or checkpoint is required. Decide what
is worth checking, when to check it, and whether to check anything at all.
\end{sglisting}

The relation schema, lifecycle transitions, tool definitions, and repair protocol remain unchanged.
Thus, Free-Form Memory preserves the machinery for persistent memory and intervention while removing
the structured policy that specifies how analytical state should be externalized and validated.

\begin{table}[t]
\centering
\small
\caption{Controlled analytical-state ablations. ``Free'' denotes an available capability without
StateGuard's prescribed state or verification policy.}
\label{tab:ablation_configurations}
\begin{tabular}{lcccccc}
\toprule
Variant & Lifecycle & State & Tools & Repair & Memory & State/Check Policy \\
\midrule
ReAct
& -- & -- & -- & -- & -- & -- \\
w/o State Validity
& $\times$ & $\times$ & $\times$ & $\times$ & $\times$ & $\times$ \\
Free-Form Memory
& $\checkmark$ & $\checkmark$ & $\checkmark$ & $\checkmark$ & $\checkmark$ & Free \\
StateGuard
& $\checkmark$ & $\checkmark$ & $\checkmark$ & $\checkmark$ & $\checkmark$ & Structured \\
\bottomrule
\end{tabular}
\end{table}

Both state ablations use the stock Qwen3-8B manager rather than the StateGuard SFT/RL checkpoint,
preventing learned state-maintenance behavior from leaking into the ablated settings. The resulting
progression separates three effects: local observation alone, persistent but unrestricted state
management, and the structured state-validity mechanism used by StateGuard.

\section{Additional Analyses and Case Studies}

\subsection{Runtime Overhead}
\label{app:runtime_overhead}

StateGuard introduces additional computation through periodic manager activations. Since raw
end-to-end wall time is jointly affected by worker budget, task length, and concurrent service
contention, we estimate manager overhead from the execution trace rather than attributing the
entire wall-time difference to StateGuard.

\paragraph{Measurement Protocol.}
Worker and manager execution are serialized: at each review boundary, the worker pauses until the
manager cycle completes. Let
\[
\Delta_{\mathrm{bdry}}
\quad\text{and}\quad
\Delta_{\mathrm{local}}
\]
denote the average worker-step interval at review boundaries and ordinary non-boundary positions,
respectively. We estimate the incremental cost of one manager cycle as
\begin{equation}
\label{eq:manager_overhead}
\widehat{T}_{M}
=
\Delta_{\mathrm{bdry}}
-
\Delta_{\mathrm{local}}
-
\Delta_{\mathrm{harness}},
\end{equation}
where $\Delta_{\mathrm{harness}}$ is the corresponding boundary bookkeeping cost measured from the
manager-free run. The per-task manager cost is then
\[
\widehat{T}_{M}^{\mathrm{task}}
=
N_{\mathrm{review}}
\cdot
\widehat{T}_{M}.
\]
Intervals longer than 900 seconds are treated as interrupted sessions and excluded from the timing
decomposition.

\paragraph{Observed Cost.}
Table~\ref{tab:runtime_overhead} summarizes the resulting manager-side load. The average latency of
an individual manager-model call remains on the order of tens of seconds across all benchmarks.
The principal determinant of total overhead is therefore review frequency rather than repair
frequency.

\begin{table}[h]
\centering
\small
\caption{Runtime characteristics of StateGuard with DeepSeek-V4-Pro workers. Manager inference is
served locally with Qwen3-8B.}
\label{tab:runtime_overhead}
\begin{tabular}{lccc}
\toprule
Benchmark
& Reviews / Task
& Manager Calls / Task
& Time / Manager Call \\
\midrule
LongDS-Bench & 33.1 & 134.5 & 18.4 s \\
DABstep      & 2.3  & 7.3   & 15.0 s \\
DAComp-DE (Impl.) & 12.0 & 30.6 & 31.5 s \\
DAComp-DE (Evol.) & 12.3 & 35.1 & 26.9 s \\
\bottomrule
\end{tabular}
\end{table}

The scaling behavior is consistent with the runtime design: benchmarks with denser review
boundaries invoke the manager more frequently, while sparse-cadence settings incur substantially
fewer manager cycles. This makes the additional computation predictable from the review schedule
and allows the accuracy--latency trade-off to be adjusted directly through the cadence parameter.

\paragraph{Repair and Monetary Cost.}
Most manager activity is devoted to state maintenance and verification rather than intervention.
The average number of \textsc{Repair} actions is only 0.20--0.52 per task across the evaluated
settings, indicating that repair is selectively triggered rather than used as a recurrent
regeneration mechanism.

All manager inference is served locally with the Qwen3-8B StateGuard model. Consequently,
manager activations introduce GPU computation but no additional external-model API calls. Any
increase in worker-side API usage arises only when the worker is given additional execution steps
to respond to manager feedback. Thus, StateGuard primarily trades local inference compute for
persistent state maintenance and verification, with the overall cost controlled by the chosen
review cadence.

\subsection{State and Intervention Statistics}
\label{app:state_intervention_stats}

We further inspect the internal behavior of StateGuard when paired with the DeepSeek-V4-Pro worker.
The statistics reveal a consistent pattern: StateGuard maintains persistent analytical context
through structured states and relations, while explicit intervention remains sparse and selective.

\begin{figure}[h]
    \centering
    \begin{subfigure}[t]{0.49\linewidth}
        \centering
        \includegraphics[width=\linewidth]{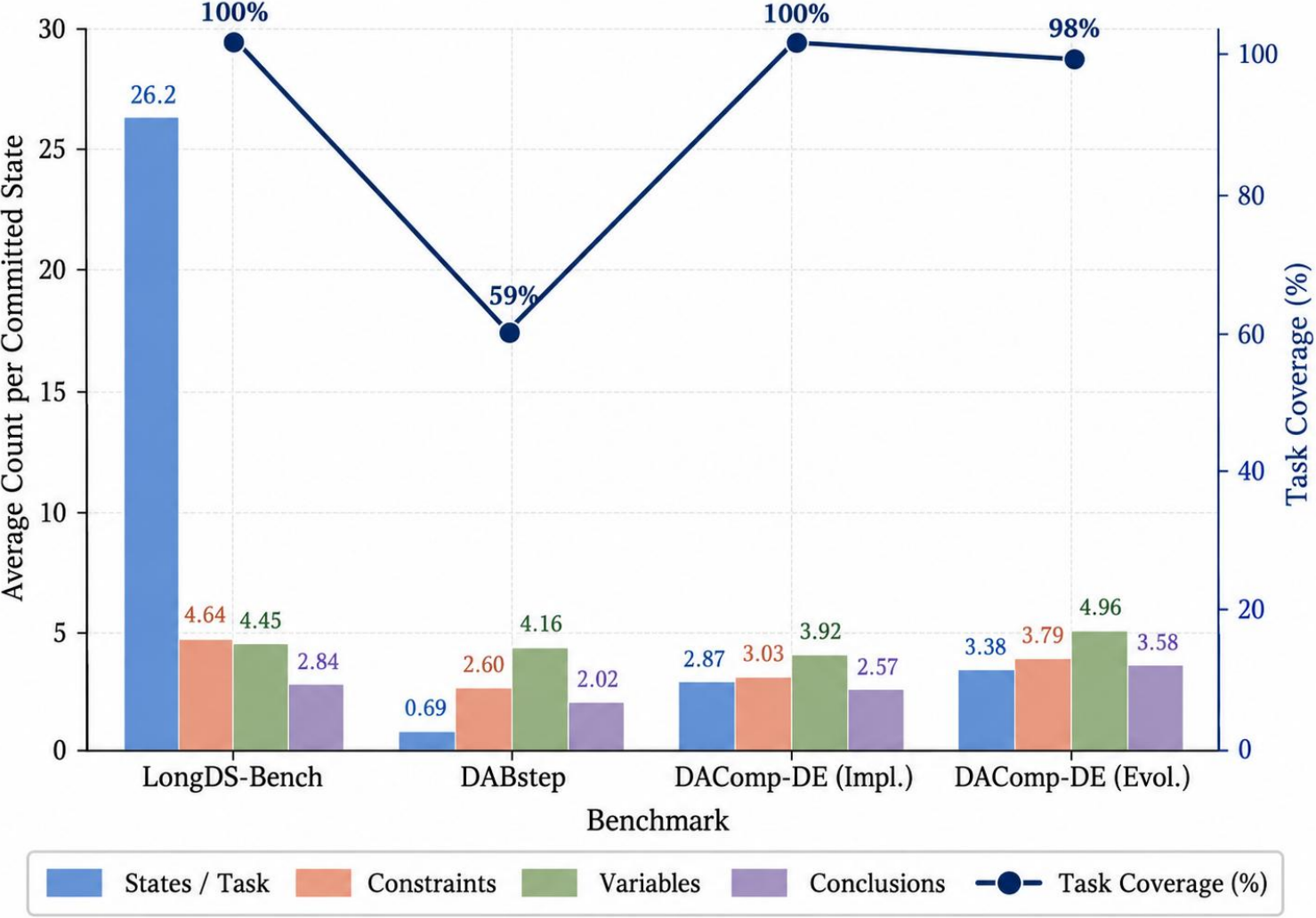}
        \caption{Analytical-state statistics.}
        \label{fig:state_stats}
    \end{subfigure}
    \hfill
    \begin{subfigure}[t]{0.49\linewidth}
        \centering
        \includegraphics[width=\linewidth]{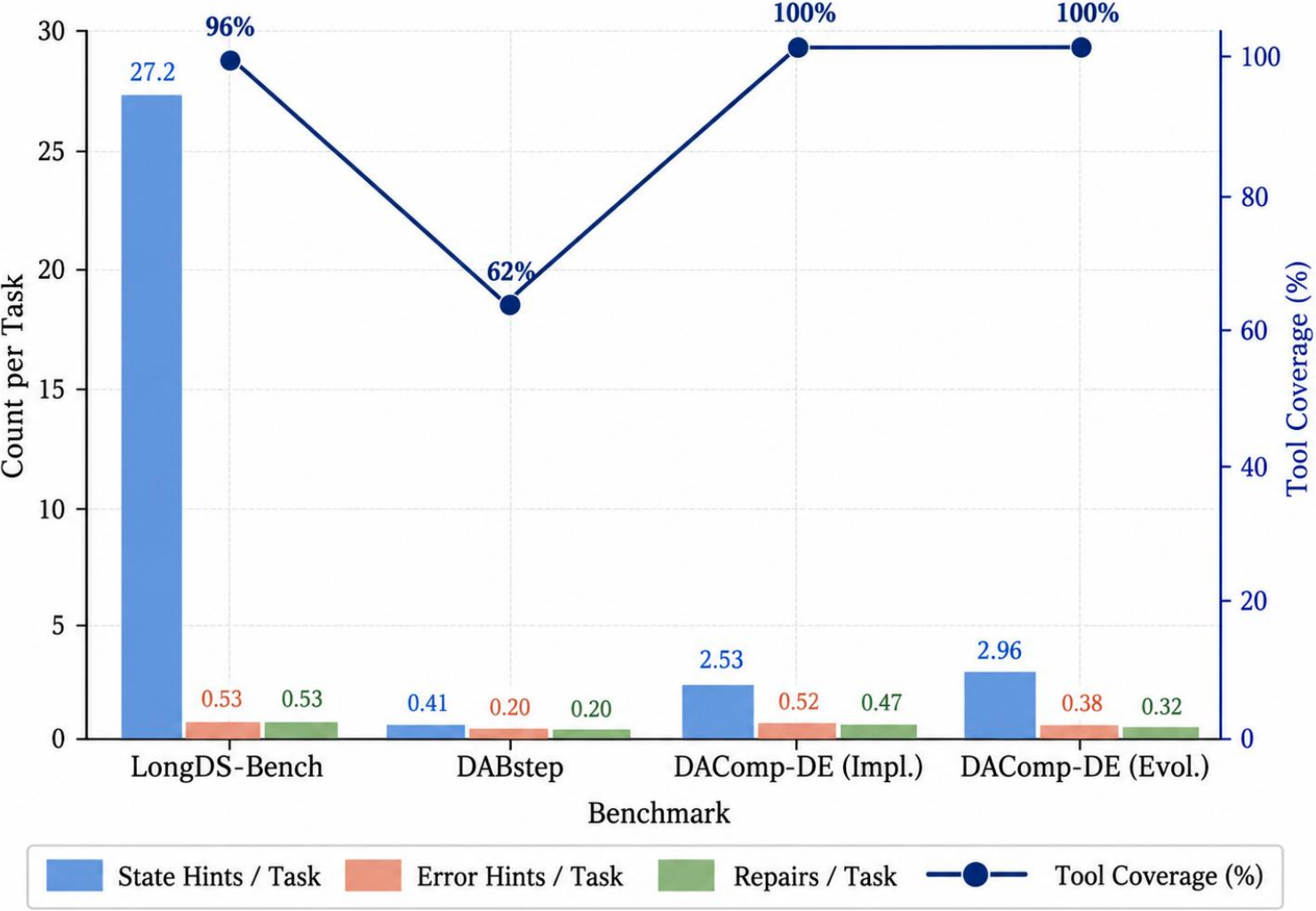}
        \caption{Intervention and verification statistics.}
        \label{fig:intervention_stats}
    \end{subfigure}
    \caption{Internal runtime statistics of StateGuard with DeepSeek-V4-Pro as the worker.
    Left: state construction statistics, including average states per task, task coverage, and
    per-state content density. Right: intervention behavior, including the frequency of state hints,
    error hints, and repairs, as well as tool-usage coverage. Across benchmarks, StateGuard maintains
    dense structured analytical states while keeping explicit repair interventions sparse and selective.}
    \label{fig:state_intervention_stats}
\end{figure}

\paragraph{State Construction.}
As shown in Figure~\ref{fig:state_stats}, StateGuard materializes non-trivial analytical states
throughout execution. LongDS produces 26.2 committed states per task; equivalently, 1,785 of the
2,225 evaluated turns (80.2\%) receive their own committed analytical state. State construction is
also nearly universal on DAComp-DE, covering all Implementation tasks and 98\% of Evolution tasks.

The resulting states contain substantive analytical content rather than lightweight bookkeeping records.
Across benchmarks, a committed state typically contains 3--4 constraints, 3--5 variables, and
2--3 intermediate conclusions. Moreover, most recorded variables carry concrete values, indicating
that the state store preserves executable analytical artifacts rather than only textual summaries.

\paragraph{State Relations.}
The learned relation structure reflects the underlying workflow rather than collapsing to a single
transition pattern. LongDS exhibits the richest dependency structure: among 2,399 committed
relations, 44.0\% are \textsc{Combine}, 34.3\% are \textsc{Progress}, and 14.6\% are \textsc{Branch};
441 states have multiple upstream relations. This matches its multi-turn analytical setting, where
later questions frequently synthesize results from several earlier turns. By contrast, DAComp-DE is
predominantly sequential, with \textsc{Progress} accounting for 56.8\% and 64.5\% of relations in
Implementation and Evolution, respectively.

The two lifecycle regimes also behave as intended. In turn-aligned LongDS, relations are proposed
when a state is opened and subsequently finalized; only 25 cases require reselection. For fixed-cadence
workflows, relations are selected post hoc at finalization, matching the segment-induced design
described in Appendix~\ref{app:lifecycle_relation}.

\paragraph{Hints, Verification, and Repair.}
Figure~\ref{fig:intervention_stats} shows that StateGuard primarily communicates persistent analytical
context rather than frequent corrective interventions. In LongDS, 1,827 of 1,863 delivered hints are
state hints, compared with only 36 error hints. The same pattern holds on DAComp-DE, where state
summaries substantially outnumber error hints. Thus, the dominant runtime behavior is continuous state
propagation rather than repeated error correction.

Verification is nevertheless widely used: evidence tools are invoked on 96\% of LongDS tasks and all
DAComp-DE tasks. The manager mainly relies on committed-state retrieval, execution checking, syntax
validation, and targeted probes to establish local evidence before acting. Repair is deliberately
conservative: across all four settings, only 74 repair actions are issued, corresponding to
0.20--0.52 repairs per task, and nearly all occur as a first repair attempt. Heavy repair is less required. This is consistent with StateGuard's design: the manager continuously maintains and
verifies analytical state, but intervenes only when localized evidence provides sufficient support for
correction.

\subsection{Additional Case Analysis}
\label{app:case_analysis}
We further conduct paired case analysis with DeepSeek-V4-Pro as the worker. Fine-grained pairing
is available for LongDS at the turn level and DAComp-DE at the task level; DABstep is excluded
because its full test-set ground truth is withheld.

\begin{figure}[h]
    \centering
    \includegraphics[
        width=0.8\linewidth,
        height=0.55\linewidth
    ]{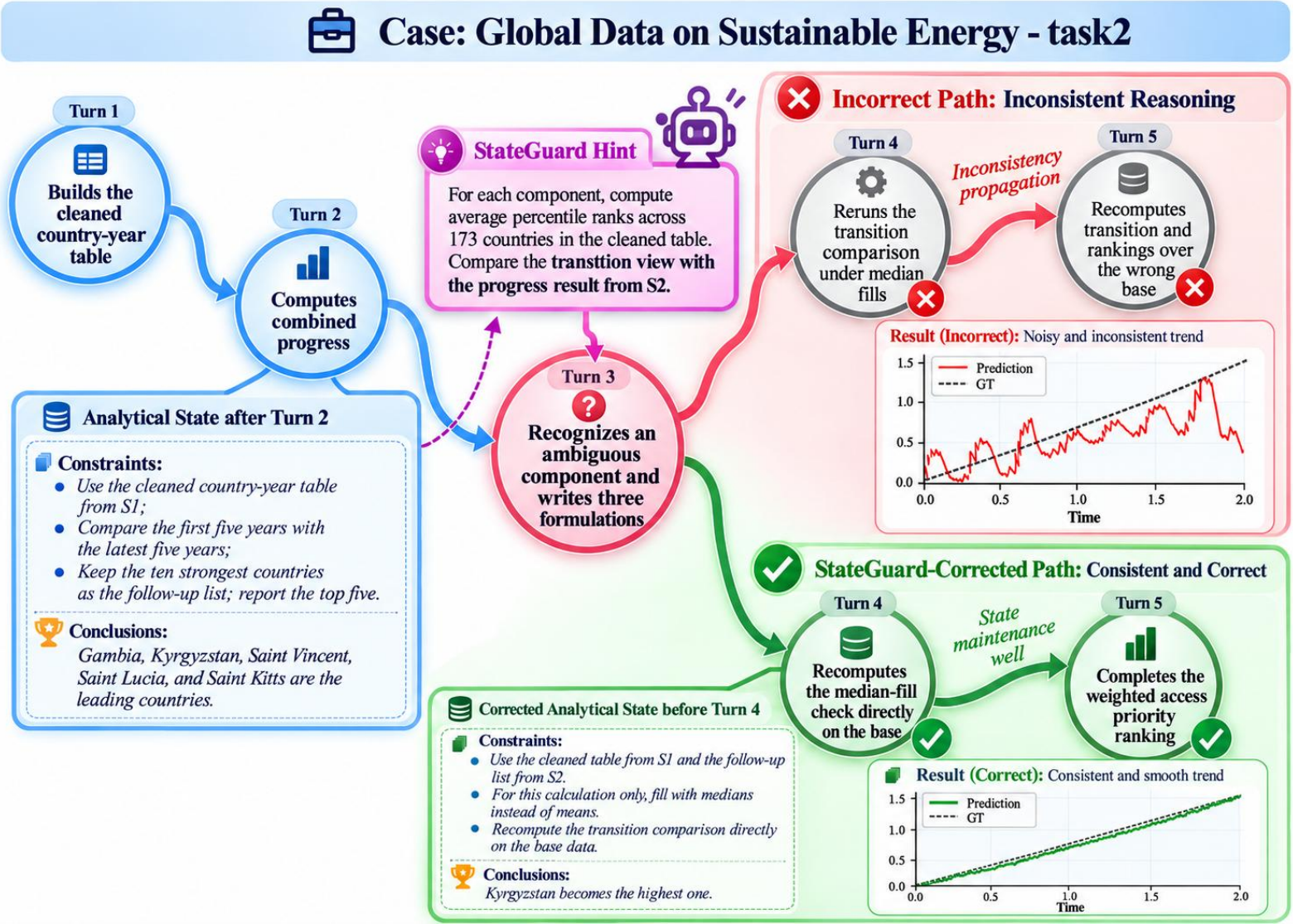}
    \caption{Representative LongDS case from \textit{Global Data on Sustainable Energy}.
    An ambiguity at Turn~3 leads to two downstream paths: the upper path propagates an inconsistent
    analytical base, whereas StateGuard retrieves the relevant upstream state, prompts localized
    recomputation, and preserves a consistent state for subsequent ranking.}
    \label{fig:case_sustainable_energy}
\end{figure}

\paragraph{Overall Paired Behavior.}
On LongDS, 350 of 2,225 turns improve from incorrect to correct, while 140 regress, yielding a net
gain of 210 correct turns. At the task level, 35 of 68 tasks improve, 23 decrease, and 10 remain
unchanged. Since raw regressions may also arise from worker stochasticity, we further identify cases
with an explicit StateGuard propagation path. Only 73 turns (3.28\%) satisfy this broader
manager-attributable criterion, and 23 turns (1.03\%) remain under a stricter criterion requiring
an incorrect state value to be directly reused downstream. On DAComp-DE, most tasks remain stable:
Implementation has 9 improvements, 4 regressions, and 17 unchanged tasks, while Evolution has
13, 5, and 32, respectively.

\textbf{Positive Case: Recovering a Consistent Analytical Path.}
Figure~\ref{fig:case_sustainable_energy} shows a representative LongDS example from \emph{Global Data on Sustainable Energy}.
An ambiguous formulation at Turn 3 causes the baseline worker to continue from an inconsistent
analytical base. StateGuard retrieves the relevant upstream state and provides a localized hint,
prompting recomputation on the intended cleaned table and preserving a consistent downstream
ranking. The intervention restores analytical context rather than supplying a replacement solution.

\textbf{Failure Case: Propagation from an Incorrect Committed State.}
A complementary failure occurs when an incorrect upstream artifact survives verification and is
committed. In a Kaggle Survey 2019 task, erroneous job-listing counts are later correctly retrieved
as dependencies and reused as denominators, propagating numerical errors downstream. This failure
highlights that validity and consistency checks cannot always detect semantically incorrect yet
internally consistent analysis, motivating stronger semantic verification.

\section{Limitations and Future Directions}
\label{app:limitations}

\paragraph{Internalizing State Management.}
StateGuard currently relies on an external manager to maintain analytical state, trace dependencies,
and regulate validity throughout long-horizon execution. This design provides explicit control and
interpretability, but still treats state management as an auxiliary mechanism around the worker.
A longer-term direction is to internalize these capabilities into the worker itself, so that state
awareness, revision tracking, dependency-sensitive reasoning, and selective validation become intrinsic
parts of the agent policy rather than externally orchestrated behaviors. Such integration may also
reduce coordination overhead and improve the agent's ability to anticipate state invalidation before
an explicit manager intervention is required.

\paragraph{Lightweight and Adaptive Management.}
Persistent state maintenance introduces additional inference cost, particularly when review points are
dense or trajectories are long. Future work could make StateGuard substantially more efficient through
smaller distilled managers, adaptive review cadence, or event-triggered activation based on state
uncertainty, dependency depth, and inconsistency risk. A hierarchical design is also promising, where
cheap deterministic checks handle routine validation and the manager is invoked only for ambiguous or
high-risk cases. This would allow StateGuard to preserve most of its state-management benefits while
reducing unnecessary manager calls and improving deployment efficiency.

\paragraph{Stronger Validity Guarantees.}
Current verification is primarily evidence-driven and localized to the affected analytical state.
Although this is effective for detecting many execution and consistency errors, it does not provide
formal guarantees over the full analytical dependency graph. Future work could incorporate stronger
executable invariants, typed state representations, provenance-aware dependency checking, and global
cross-state consistency constraints. Combining deterministic verification with learned semantic
judgment may further improve robustness, especially for failures that cannot be reduced to local
execution checks. Such mechanisms could reduce the chance an invalid upstream artifact is
committed, reused, and propagated through a long chain of downstream states.
\end{document}